\documentclass[sigconf,natbib=true,anonymous=false]{acmart}
\setcopyright{none}
 \renewcommand\footnotetextcopyrightpermission[1]{} 
 
\AtBeginDocument{%
  \providecommand\BibTeX{{%
    \normalfont B\kern-0.5em{\scshape i\kern-0.25em b}\kern-0.8em\TeX}}}

\usepackage{amsmath,amsfonts}
\usepackage{dblfloatfix}
\usepackage{graphicx}
\usepackage{hyperref}
\usepackage{textcomp}
\usepackage{caption}
\usepackage{subcaption}
\usepackage{xcolor}
\usepackage{comment}
\usepackage[export]{adjustbox} 
\usepackage{multirow}
\usepackage{verbatim}
\usepackage{listings}
\usepackage[normalem]{ulem}
\usepackage[ruled,vlined, linesnumbered]{algorithm2e}
\SetKwRepeat{Do}{do}{while}
\usepackage[colorinlistoftodos]{todonotes}

\newcommand{\method}{DenDrift} 

\begin{document}

\title{DenDrift: A Drift-Aware Algorithm for Host Profiling}

\author{Ali Sedaghatbaf}
\affiliation{%
  \institution{RISE Research Institutes of Sweden}
  \city{Västerås}
  \country{Sweden}
}
\email{ali.sedaghatbaf@ri.se}

\author{Sima Sinaei}
\affiliation{%
  \institution{RISE Research Institutes of Sweden}
  \city{Västerås}
  \country{Sweden}
}
\email{sima.sinaei@ri.se}

\author{Perttu Ranta-aho}
\affiliation{%
  \institution{F-Secure Corporation}
  \city{Helsinki}
  \country{Finland}}
\email{perttu.ranta-aho@f-secure.com}

\author{Marko Koskinen}
\affiliation{%
  \institution{Techila Technologies}
  \city{Tampere}
  \country{Finland}
}
\email{marko.koskinen@techilatechnologies.com}

\renewcommand{\shortauthors}{Sedaghatbaf and Sinaei, et al.}

\begin{abstract}
Detecting and reacting to unauthorized actions is an essential task in security monitoring. What make this task challenging are the large number and various categories of hosts and processes to monitor. To these we should add the lack of an exact definition of normal behavior for each category. Host profiling using stream clustering algorithms is an effective means of analyzing hosts' behaviors, categorizing them, and identifying atypical ones. However, unforeseen changes in behavioral data (i.e. concept drift) make the obtained profiles unreliable. DenStream is a well-known stream clustering algorithm, which can be effectively used for host profiling. This algorithm is an incremental extension of DBSCAN which is a non-parametric algorithm widely used in real-world clustering applications. Recent experimental studies indicate that DenStream is not robust against concept drift. In this paper, we present \method{}  as a drift-aware host profiling algorithm based on DenStream. \method{} relies on non-negative matrix factorization for dimensionality reduction and Page-Hinckley test for drift detection. We have done experiments on both synthetic and industrial datasets and the results affirm the robustness of \method{}  against abrupt, gradual and incremental drifts.  
\end{abstract}





\maketitle
\pagestyle{plain} 
\section{Introduction}\label{sec:intro}
F-Secure is known as a reliable cyber-security leader around the world. F-Secure has over 100,000 corporate customers, serves more than 300 enterprises through consulting, and tens of millions of consumer customers are protected via the security products provided by this corporation. Among other tasks, the detection and response system at F-Secure collects process execution data from a set of hosts monitored by this system. The collected data indicate when and how many times each process is executed by each host. Currently an offline clustering algorithm is used to group hosts based on their similarities in process execution. Anomaly detection based on offline clustering is challenging due to the statistical changes in data which are known as \textit{concept drift} (see Section \ref{sec:drift}). Concept drift has a negative impact on the quality of clusters and imposes further effort for anomaly detection. In particular, the clusters obtained for one point of time might be much different from the previous point due to statistical changes in data over time. 
F-Secure is looking for a clustering algorithm for host profiling that can detect and adapt to concept drifts. No presumption should be made about the number of clusters, and the algorithm should be able to produce clusters of arbitrary shapes. 

Our proposition is to use an incremental and non-parametric clustering algorithm that can maintain a high clustering quality by properly detecting concept drift in the input stream and retraining the underlying model after drift. In incremental clustering, efficient processing and memory consumption is of paramount importance. On one hand, data is constantly arriving and we have a limited time for data processing. On the other hand, we cannot store all data in memory since the data stream may exist for several months or years. Several incremental clustering algorithms have been developed in the past two decades. These algorithms can be categorized into the following four groups \cite{moulton2018clustering}: (1) \textit{partitioning algorithms} (e.g. CluStream \cite{aggarwal2003framework}, StreamKM++ \cite{ackermann2012streamkm++}) which partition the data stream into a predefined number of groups, (2) \textit{density-based algorithms} (e.g. DenStream \cite{cao2006density}, SDSttream \cite{ren2009density}) which try to identify high-density regions in the data stream, (3) \textit{grid-based algorithms} (e.g. DStream \cite{chen2007density}, DENGRIS \cite{amini2012dengris}) which divide the feature space into grid cells and then map each data point to a cell, and (4) \textit{hierarchical algorithms} (e.g. ClusTree \cite{kranen2011clustree})
which rely on a hierarchical index structure to maintain stream summaries. Among these algorithms, we are more interested in DenStream, which is computationally efficient and makes no limiting assumption on the shape and number of clusters.

In this paper, we introduce \textit{\method{}}  as an algorithm for host profiling under concept drift. As input, \method{}  receives a stream of process execution events. Each event in this stream records the number of process executions per host. 
Since we have thousands of hosts and processes, clustering hosts directly based on the host-process events is computationally intractable, \method{}  generates a host-process matrix for each time interval (with a user-defined duration), and uses non-negative matrix factorization (NMF) \cite{lee1999learning} to reduce the dimensionality of that matrix. In this way, the stream of execution events is transformed to a stream of matrices. For clustering this stream, \method{}  relies on DenStream. DenStream is simple, efficient, can generate clusters of arbitrary shape, can effectively detect anomalies, and has demonstrated a good clustering quality. However, experimental studies \cite{moulton2018clustering} indicate that this algorithm is not robust against concept drift. In particular, concept drift leads to a decrease in the quality of clusters generated by this algorithm. To overcome this limitation, \method{}  uses the Page-Hinckley Test (PHT) \cite{mouss2004test} for drift detection, and retrains the DenStream model for drift adaptation.

Our experimental results (see Section \ref{sec:exp}) affirm that not only \method{}  improves the robustness of DenStream against three prevalent forms of concept drift (i.e. abrupt, gradual and incremental), but also it can effectively distinguish abrupt drifts from outliers. In this paper, we report experiments on both synthetic and real-world data. The real-world data are obtained from the event log files recorded by the detection and response system at F-Secure. To generate synthetic data with controlled concept drift, we have developed a drifted stream generator using Gaussian mixture models. In summary, the contributions of this paper include the following:
\begin{itemize}
    \item Elaborating \method{}   as a new algorithm for host profiling, which is robust against abrupt, gradual and incremental drifts and can detect outliers,
    \item Introducing a drifted stream generator based on Gaussian mixture models that can simulate concept drift and concept evolution,
    \item Explaining the experiments performed on synthetic and industrial datasets, and
    \item Discussing about the computational efficiency of \method{}.
\end{itemize}

The rest of the paper is organized as follows. Section \ref{sec:related} is dedicated to discussion about the related work. Some background information about DenStream and concept drift is provided in Section \ref{sec:backgrnd}. An overview of \method{}  is presented in Section \ref{sec:method}. The implementation and experiments are detailed in Section \ref{sec:exp}, and finally some concluding remarks and future directions are pointed out in Section \ref{sec:conclusion}.

\section{Related Work}\label{sec:related}


The research contributions related to the scope of this paper can be categorized into two groups: (1) algorithms for dimensionality reduction, and (2) concept drift detection methods.
\subsection{Dimensionality Reduction}\label{sec:rel_dim}
Analyzing high dimensional data is a significant challenge in machine learning. On one hand, recent advances in data collection tools and techniques have led to the generation of huge amounts of data with several dimensions (or features). On the other hand, not all features may be relevant and useful to extract valuable information. Several Dimensionality Reduction (DR) algorithms \cite{ayesha2020overview} have been proposed to eliminate irrelevant features and improve the efficiency and accuracy of the machine learning algorithms processing the collected data. 

Principal Component Analysis (PCA) \cite{Jolliffe2011} is a well-known and unsupervised DR algorithm whose goal is to discover a set of features referred to as principal components. These features hold the highest variance hence the maximum information. PCA has been applied to several domains (e.g. image and voice processing), and several extensions have been introduced for this algorithm since introduction. Single Value Decomposition (SVD) \cite{modarresi2015unsupervised} is another DR algorithm closely related to PCA, and specifically developed for matrix factorization. This algorithm transforms the input matrix to two orthogonal and one diagonal matrices. Latent Semantic Analysis (LSA) \cite{tang2005comparing} is another DR technique based on SVD, which is typically used for extracting information from text documents.

In this paper we use NMF \cite{lee1999learning} for reducing the dimensionality of host-process matrices. NMF is an algorithm for factorizing matrices with non-negative elements, and has been widely applied in several fields such as topic discovery \cite{shi2018short}, age prediction \cite{varikuti2018evaluation}, image clustering \cite{yang2018non} and speech enhancement \cite{bando2018statistical}.We prefer NMF to PCA and SVD due to several reasons. It is more suitable for very sparse matrices (as in our case), and we  do not need to make any assumption about the missing values. Furthermore, through non-negativity constraints, NMF allows only additive combinations, which makes it capable of learning parts-based representations. All the values recorded in the event log provided by F-Secure are non-negative. Therefore NMF is suitable for reducing the dimensionality of host-process matrices. Last but not the least, NMF has an inherent clustering property \cite{ding2005equivalence} which make it capable of discovering natural clusters and mislabeled records while PCA and SVD cannot. 

\subsection{Concept Drift Detection}\label{sec:rel_drift}
Concept drift detection is essential to trigger stream adaptation strategies in data stream mining. Several drift detection methods have been proposed recently \cite{gonccalves2014comparative}. The techniques can be divided into two categories based on their reliance on labeled data: explicit/supervised drift detectors and implicit/unsupervised drift detectors. Explicit drift detectors, which focus on \textit{real drifts} (as defined in Section \ref{sec:drift}) rely on labeled data to compute performance metrics such as accuracy and F-measure, which they can monitor online over time. They detect a drop in performance and as such, are efficient in signaling drift when it occurs. Implicit drift detectors focus on \textit{virtual drifts} (as defined in Section \ref{sec:drift}) and rely on properties of the unlabeled data’s feature values to signal deviations. They are prone to false alarms, but their ability to function without labeling makes them useful in applications where labeling is expensive, time-consuming, or not available. Our main focus in this paper is on virtual drifts and unsupervised drift detectors applied to clustering problems.

The OnLIne Drift Detection Algorithm (OLINDDA) uses K-means clustering to continuously monitor and adapt to emerging data distributions \cite{spinosa2007olindda}. Unknown samples are stored in a short-term memory queue and are periodically clustered and then either merged with existing similar clusters or added as a new cluster to the pool of clusters. When a new cluster forms, it will be considered either a novel concept or a drifted concept depending on the distance between its centroid and the global centroid (i.e. the centroid of centrods). If the distance is less than a given threshold, it's a drifted concept and a novel concept otherwise. 
The MINAS algorithm \cite{faria2013novelty} is another distance-based method for drift and novelty detection. This algorithm relies on CluStream for stream clustering. CluStream is an incremental extension of K-means. Instead of the distance from the global concept, the distance between the centroid of the newly-formed cluster and the centroid of its nearest cluster is investigated by MINAS for drift and novelty detection. Short distances indicate drift and long ones imply novelty. 

In \cite{zgraja2018drifted}, the authors propose a solution to improve the robustness of ClusTree \cite{kranen2011clustree} against concept drift. ClusTree is a hierarchical stream clustering algorithm that relies on R-trees to maintain a stream summary. The solution proposed in \cite{zgraja2018drifted} relies on the sliding window model and the number of novelties for concept drift detection. In particular, if the new data point doe not belong to any of the micro-clusters live in the current window, it is recognized as a novelty, and we have a concept drift if the quantity of novelties is higher than a given threshold. DriftVis \cite{yang2020diagnosing} is another solution based on sliding windows and distance measurement. In this method, the energy distance between the new data point and the most similar ones is calculated as a measure of concept drift. Gaussian mixture models are used to cluster data over time, and the points (in the latest window) that go to the same cluster with the new data point, are considered most similar to it. A high energy distance is then considered as an indicator of concept drift. 

While most of the related work have focused on abrupt drifts (e.g. \cite{spinosa2007olindda, faria2013novelty, zgraja2018drifted, yang2018non}), \method{}  can detect also incremental and gradual drifts. \method{}  relies on PHT \cite{mouss2004test} for drift detection. PHT is a simple and efficient means of change detection. Furthermore, unlike OLINDDA, MINAS and the solution proposed in \cite{zgraja2018drifted}, drift detection in \method{} is independent of the underlying clustering algorithm, which makes \method{}  extensible to other incremental clustering algorithms (e.g. CluStream) as well.

\section{Background}\label{sec:backgrnd}
\subsection{Concept Drift}\label{sec:drift}
Assume that we have a stream of data points such that each point $d$ in this stream is defined as a pair $(x,y)$ where $x$ is the feature vector and $y$ is the class label of $d$. Any change in the statistical properties of $x$ and/or the probability $p(y|x)$ is recognized as a \textit{concept drift}. If the change occurs only in $p(x)$ without affecting $p(y|x)$, then the concept drift is called \textit{virtual}. On the other hand,  changes in $p(y|x)$ which may occur with/without changes in $p(x)$ are called \textit{real} drifts \cite{gama2014survey}. In this paper we focus on virtual drifts since there is no class label in the dataset provided by F-Secure. To detect virtual drifts we investigate changes in data \textit{mean} over time. 

\begin{figure}[tb]
     \centering
     \begin{subfigure}[b]{0.3\textwidth}
         \centering
         \includegraphics[width=\textwidth]{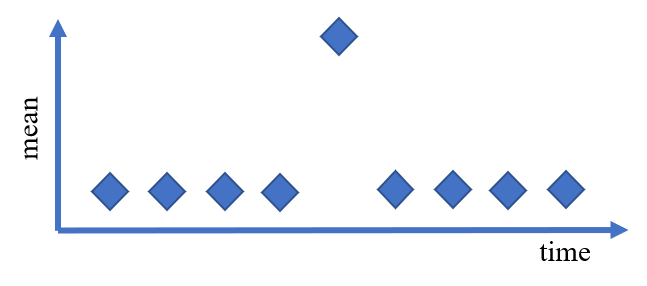}
         \caption{outlier}
         \label{fig:outlier}
     \end{subfigure}
     \hfill
     \begin{subfigure}[b]{0.3\textwidth}
         \centering
         \includegraphics[width=\textwidth]{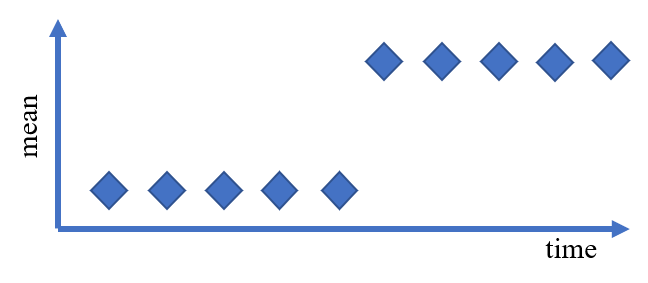}
         \caption{abrupt drift}
         \label{fig:ab_drift}
     \end{subfigure}
     \hfill
     \begin{subfigure}[b]{0.3\textwidth}
         \centering
         \includegraphics[width=\textwidth]{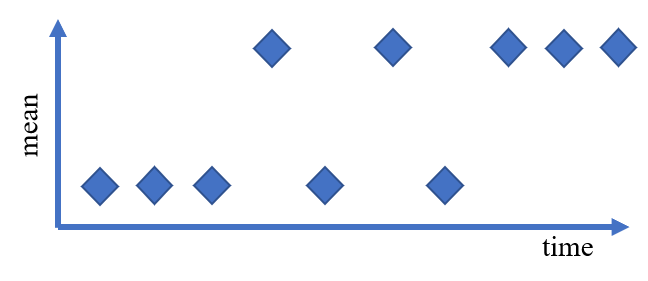}
         \caption{gradual drift}
         \label{fig:gr_drift}
     \end{subfigure}
     \hfill
     \begin{subfigure}[b]{0.3\textwidth}
         \centering
         \includegraphics[width=\textwidth]{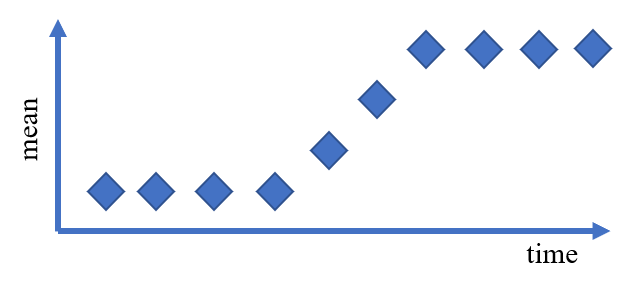}
         \caption{incremental drift}
         \label{fig:syn_i_det}
     \end{subfigure}
     \caption{Three prevalent forms of concept drift}
     \label{fig:drift_forms}
\end{figure}

Virtual drifts may appear in several forms \cite{gama2014survey, lemaire2014survey}. In this paper we focus on three prevalent ones i.e. abrupt/sudden, gradual, and incremental (see Figure \ref{fig:drift_forms}). In abrupt drift, we see a sudden change in the mean of data points. In other words, a new stable concept replaces the current concept immediately at some point of time. In gradual drift, we see some alternations between the two concepts for a short period of time, and finally some intermediate and temporary concepts appear between the stable concepts in incremental drift. 

\subsection{DenStream}\label{sec:denstream}
DenStream \cite{cao2006density} is an incremental extension of DBSCAN \cite{ester1996density}. DBSCAN is a well-known and density-based algorithm for offline clustering. In the online learning phase, DenStrean arranges the incoming records into micro-clusters such that each micro-cluster includes records whose density is more than the given threshold $\mu$. In particular, upon the arrival of data record $R$, DenStream tries to merge it with the nearest micro-cluster $MC$. $R$ can be merged into $MC$ only if the radius of $MC$ after merging does not exceed the threshold specified by parameter $\epsilon$. If there is no micro-cluster suitable for merging, then a new micro-cluster will be created for $R$. 
To each micro-cluster a weight $w$ is assigned which decreases over time if the micro-cluster does not change. The forgetting rate is specified by parameter $\lambda$. Micro-clusters are checked periodically and categorized into two groups based on the value of $w$. The micro-clusters with a high $w$ form the group of potential micro-clusters. The weight of each potential micro-cluster is controlled by parameters $\beta$ and $\mu$ such that the least weight is $\beta \times \mu$. Upon each request for offline clustering, the potential micro-clusters will be analyzed by DBSCAN to build the final clusters. On the other hand, the low-weight micro-clusters will be considered as outliers and will be removed from the set of final clusters. The lower limit of $w$ for a micro-cluster to be considered as a potential micro-cluster is computed as a function of the checking point and the point of time when the micro-cluster was created.

\section{\method{}}\label{sec:method}
In this section, we elaborate \method{}  as a drift-aware host profiling algorithm. We assume that concept drift has a negative impact on the quality of clustering. In fact, if the incoming data are from a different distribution than the old data, then we expect a degradation in the quality of the generated clusters. Accordingly, we need to continuously monitor the changes in the input data and retrain the underlying cluster model (i.e. reinitialize the set of micro-clusters) if we have more changes than a given threshold. 
In \method{}  we use DenStream as the stream clustering algorithm and employ PHT and MNF for change detection and dimensionality reduction respectively. As input, \method{}  needs the streams's dimensionality i.e. number of host latent features ($N_f$), change threshold expected by PHT ($Th_c$), and the minimum number of changed hosts to be considered as a potential concept drift ($Th_d$). \method{}  works in two modes: \textit{normal} mode where the number of changed hosts is lower than $Th_d$, and \textit{change} mode where the minimum number of changed hosts is $Th_d$.
\method{}  consists of the following steps presented in Algorithm \ref{alg:\method{}}:

\begin{algorithm}[tb]
\SetAlgoLined
\KwIn{$N_f, Th_c, Th_d$}
E = EventLog.getEvents($t$)\;
M = generateMatrix(E)\;
[W, H] = NMF(M, $N_f$)\;
C = 0\;
\eIf{Mode is ''Normal''}{
    \ForEach{$h \in H$}{
        PrimaryPHT.add(h)\;
        \If{PrimaryPHT.detectedChange($Th_c$)}{
            C += 1\;
        }
    }
    \eIf{$C \geq Th_d$}{
        Mode = ''Change''\;
    }{
        \ForEach{$h \in H$}{
            SparePHT.add(h)\;
        }
    }
}{
    \ForEach{$h \in H$}{
        PrimaryPHT.add(h)\;
        \If{PrimaryPHT.detectedChange($Th_c$)}{
            C += 1\;
        }
    }
    \If{$C < Th_d$}{
        Mode = ''Normal''\;
        \ForEach{$h \in H$}{
            SparePHT.add(h)\;
            \If{SparePHT.detectedChange($Th_c$)}{
                C += 1\;
            }
        }
    }
    \If{$C \geq Th_d$}{ 
        DenStream.reset()\;
        PrimaryPHT.reset()\;
        SparePHT.reset()\;
    }
}
DenStream.merge(H)\;
\If{$t \% T_p = 0$}{
    DenStream.prunMicroClusters($t$)\;
}
\If{clusteringRequested()}{
DenStream.generateFinalClusters()\;
}
 \caption{\method{}}
 \label{alg:\method{}}
\end{algorithm}

\begin{enumerate}
\item In the first step (lines 1-3), \method{}  generates a host-process matrix based on the execution counts recorded in the event log on date $t$ for each pair of hosts and processes. Line 3 estimates the latent matrix of hosts by applying NMF to the matrix obtained in line 2.  Given a matrix $M=[m_{ij}] \in {\mathbb{R}}^{h \times p}$, NMF generates two matrices $H=[h_{ij}] \in {\mathbb{R}}^{h \times k} $ and $W=[w_{ij}] \in {\mathbb{R}}^{k \times p} $ such that $M \approx H \times W$. In other words, NMF tries to minimize the following objective function.

\begin{equation}\label{eq:lc}
    O = \parallel M - HW \parallel^2_F
\end{equation}

\noindent where $\parallel . \parallel^2_F$ denotes the Frobenius norm. 

In our context, $M$ is the host-process matrix obtained on a specific date $t$ such that each $m_{ij} \in M$ is a non-negative integer value expressing the number of times host $h_i$ has executed process $p_j$ during the period $[t-1,t)$. After factorization, we are interested in matrix $H$ which encodes the latent features of hosts.
\item In the second step, first we check whether the current mode of \method{} is \textit{normal} or \textit{change}. In the normal mode, first the number of changed hosts is calculated by applying PHT to $H$ (lines 6-11). Host $h$ is then considered changed if PHT detects changes in at least one of the features of the corresponding vector in $H$. PHT can detect changes in the mean of a one-dimensional stream. Therefore, for each feature we train a distinct instance of PHT. Assume that we are interested to detect changes with a minimal magnitude of $\delta$ and allowed error of $\lambda$ in a one-dimensional stream $s = {x_1, x_2,...,x_n}$. Then, a PHT instance will alarm whenever $U_n - m_n \geq \lambda$ or $M_n - U_n \geq \lambda$, where 
$m_n = min_{0 \leq k \leq n}U_k$ and $M_n = max_{0 \leq k \leq n}U_k$ such that $U_k$ is estimated using the following Equation.

\begin{equation}\label{eq:u}
    U_k = \sum_{i=0}^n (x_i - \delta/2)
\end{equation}

In line 12, the number of changed hosts is compared with $Th_d$. If at least $Th_d$ hosts have changed, \method{}  will enter the change mode. Otherwise, a spare PHT will be updated with $H$. The spare PHT is used to detect outliers (see Figure \ref{fig:outlier}), while the primary one takes part in drift detection. Using two PHTs in parallel makes \method{}  capable of differentiating abrupt drifts from outliers. In the change mode, we continue using the primary PHT to detect the end of the change period (lines 20-25), which happens if the number of changed hosts goes below $Th_d$. In this case, \method{}  enters the normal mode (line 17) and the spare PHT compares $H$ after the change period with $H$ before that. If the spare instance does not recognize a significant change ($C < Th_d$), it means that the concept after the change period is similar to the one before that. Therefore, we have an outlier and a drift otherwise. Note that the spare PHT is updated only in the normal mode (lines 14-18), so it can effectively compare the post-change concept with the pre-change concept.  If the concepts are different (i.e. we have a concept drift), DenStream and the PHT instances will be reinitialized to learn the new concept (lines 36-38). 
\item Lines 41-47 express the typical steps in DenStream. First, the current matrix is merged with the micro-clusters (see Algorithm 1 in \cite{cao2006density}) which are categorized into potential and outlier groups. Then, $t$ is compared with $T_p$ (as the minimal time span for pruning micro-clusters), to update the micro-clusters (see Algorithm 2 in \cite{cao2006density}). Here, the set of potential/outlier micro-clusters gets updated with the current high-/low-weight micro-clusters. At the end and upon a request for offline clustering, DBSCAN will be invoked to extract the final clusters from the potential micro-clusters.
\end{enumerate}

\section{Experiments}\label{sec:exp}
In this section we assess the performance of \method{} on both synthetic datasets and datasets generated by applying NMF to the log file provided by F-Secure. Before discussing the experimental results, we provide details about the implementation of \method{}  in Section \ref{sec:impl}. Then we describe the datasets in Section \ref{sec:dataset}, and finally we demonstrate the robustness of \method{}  against concept drift in Section \ref{sec:obs} by comparing the quality of clusters generated by \method{}  with those generated by DenStream. 

\subsection{Implementation}\label{sec:impl}
We have implemented \method{}  using Python 3.8. We utilized the well-known library scikit-learn for implementing DBSCAN and NMF. For PHT we employed the implementation provided in scikit-multiflow. To have a precise control over concept drift in our experiments on synthetic datasets, we implemented a drifted-stream generator. This generator is capable of synthesizing streams with abrupt, gradual or incremental concept drifts using Gaussian mixture models and the Jensen-Shannon distance \cite{endres2003new, fuglede2004jensen}. In this generator, the stable concepts before and after a drift are modeled as mixture models with one multi-variate normal distribution for each cluster, and the magnitude of the drift is controlled via the Jensen-Shannon distance. 
The Jensen-Shannon distance is the square root of the Jensen-Shannon divergence \cite{lin1991divergence}, and takes a value in $[0, 1]$. The Jensen-Shannon divergence is based on the Kullback–Leibler divergence, which can be used to measure the distance between two probability distributions. The benefit of the Jensen-Shannon divergence over Kullback–Leibler is being symmetric and having always a finite value.  we can adjust different aspects of the drifts and clusters generated by the stream generator using the parameters listed in Table \ref{tbl:param}.

\begin{table}[tb]\caption{Input parameters of the stream generator}\label{tbl:param}
\centering
\begin{tabular}{ |p{0.5cm}|p{7cm}| } 
 \hline
 $T_d$ & drift type which could be ''abrupt'', ''gradual'' or ''incremental''. \\ 
 $D_d$ & drift duration i.e. the number of instances generated between two stable concepts. Each instance includes a constant number of host latent vectors.  \\ 
 $M_d$ & drift magnitude in terms of the Jensen-Shannon distance. \\ 
 $P_d$ & drift precision which indicates the amount of acceptable deviation between the magnitude of the generated drift and the value of $M_d$.\\
 $C_b$ & number of clusters before drift.\\
 $C_a$ & number of clusters after drift.\\
 $N_f$ & number of host latent features.\\
 $S_i$ & size of each instance which indicates the number of host latent vectors included in each instance.\\
 $N_b$ & number of instances before drift.\\
 \hline
\end{tabular}
\end{table}


\begin{algorithm}[b]
\SetAlgoLined
\KwIn{$C_b, C_a, M_d, P_d$}
\KwOut{$Mix_{pre}, Mix_{pst}$}
$Mix_{pre}$ = genMixModel($C_b$)\;
\While{true}{
    $Mix_{pst}$ = genMixModel($C_a$)\;
    Distance = JensenShannon($Mix_{pre}, Mix_{pst}$)\;
    \If{$Distance = M_d \pm P_d$}{
        break\;
    }
}
 \caption{Mixture Model Generation Algorithm}
 \label{alg:mdl_gen}
\end{algorithm}

Algorithm \ref{alg:mdl_gen} presents the steps of generating Gaussian mixture models for stable concepts before and after the simulated drift. Method $genMixModel$ in this algorithm receives the number of clusters as input and samples from a multivariate normal distribution for each cluster. Then the parameters of the underlying mixture model are estimated using a goodness-of-fit test. 
After estimating the mixture model characterizing the pre-drift concept, we repeatedly generate a new model for the post-drift concept until the distance between the two models has an acceptable deviation from the expected drift magnitude.

\begin{figure*}[tb]
     \centering
     \begin{subfigure}{0.24\textwidth}
         \centering
         \includegraphics[width=\textwidth]{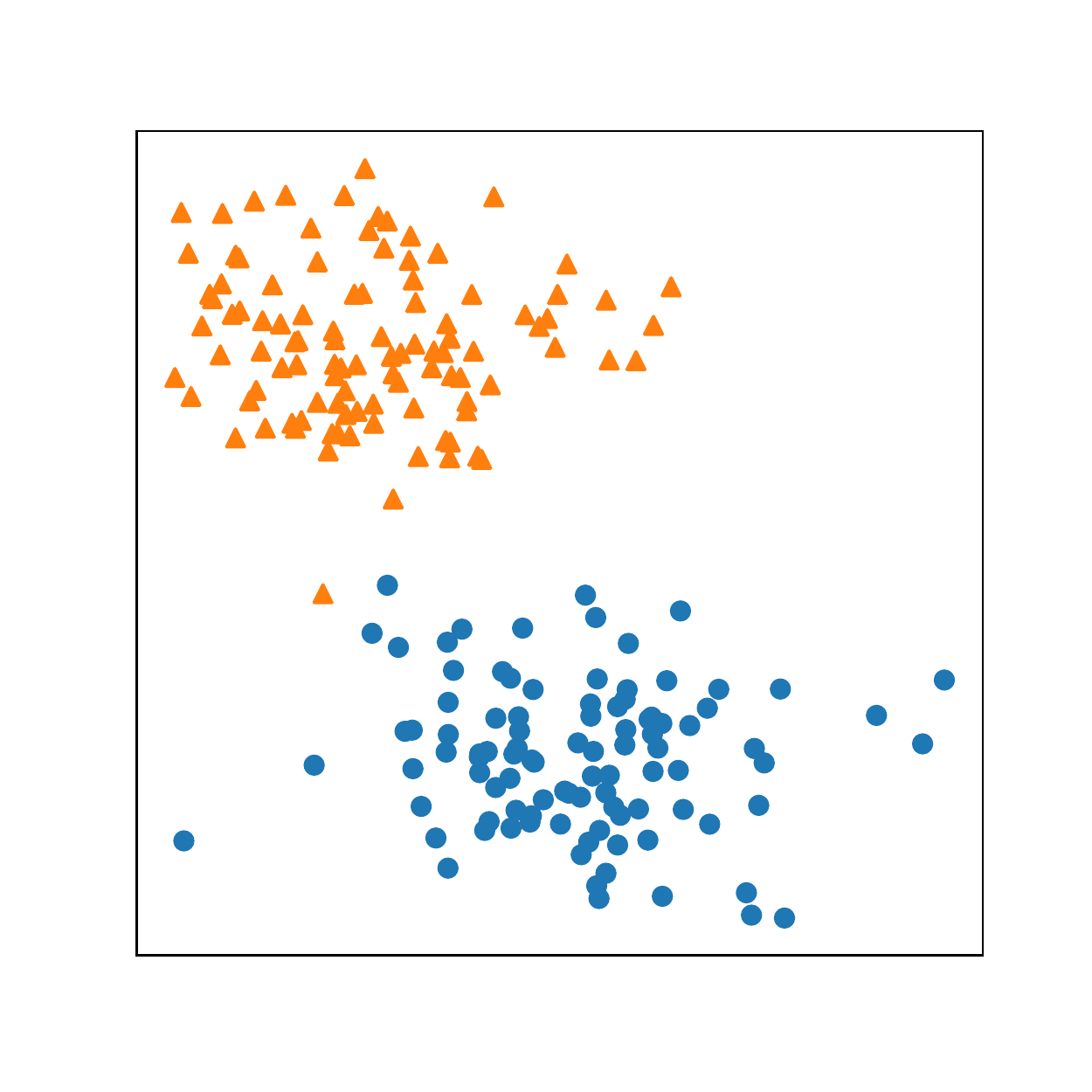}
         \caption{before drift ($C_b = 2$)}
         \label{fig:bef_drft}
     \end{subfigure}
     \hfill
     \begin{subfigure}{0.24\textwidth}
         \centering
         \includegraphics[width=\textwidth]{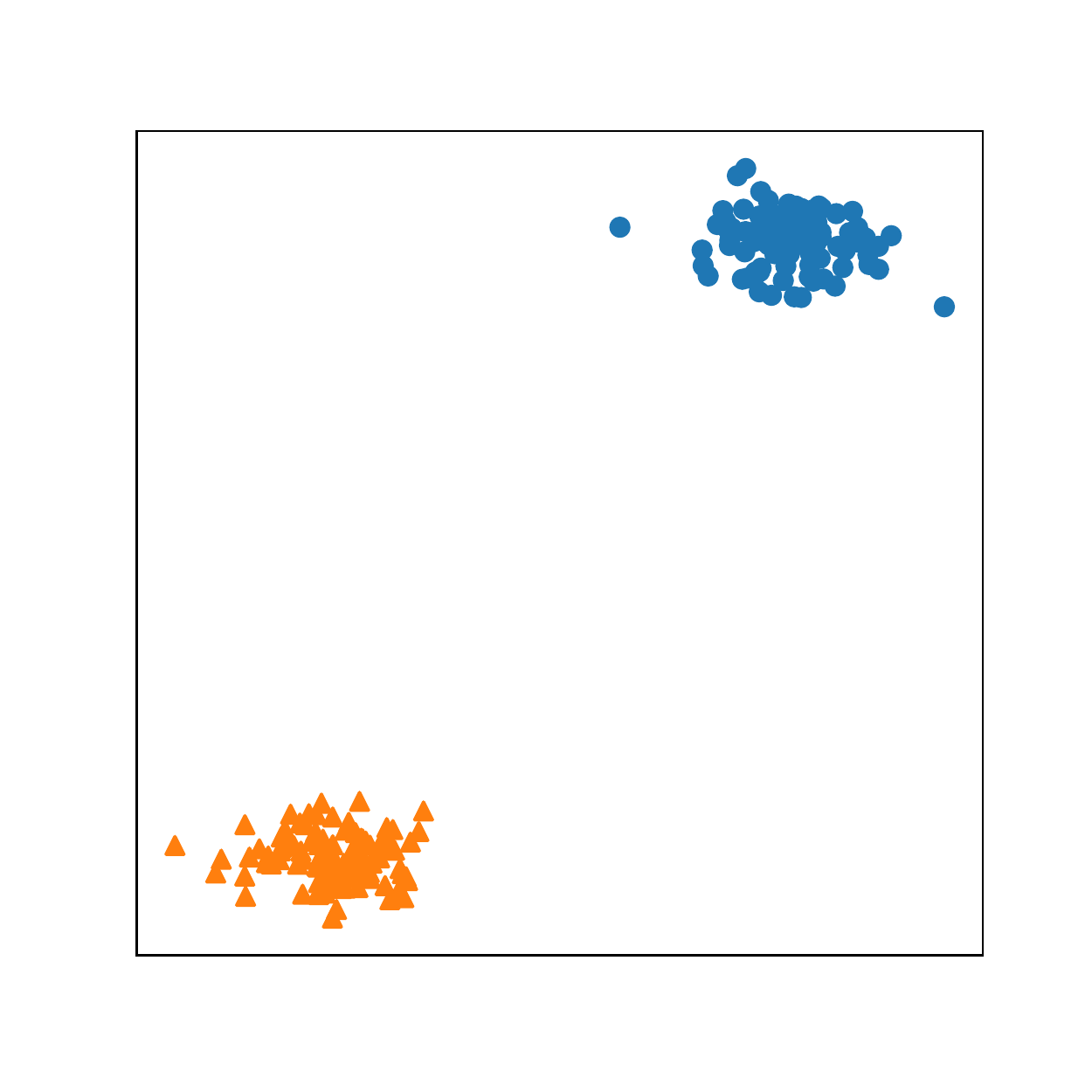}
         \caption{after drift ($M_d = 0.6, C_a = 2$)}
         \label{fig:aft_drft}
     \end{subfigure}
     \hfill
     \begin{subfigure}{0.24\textwidth}
         \centering
         \includegraphics[width=\textwidth]{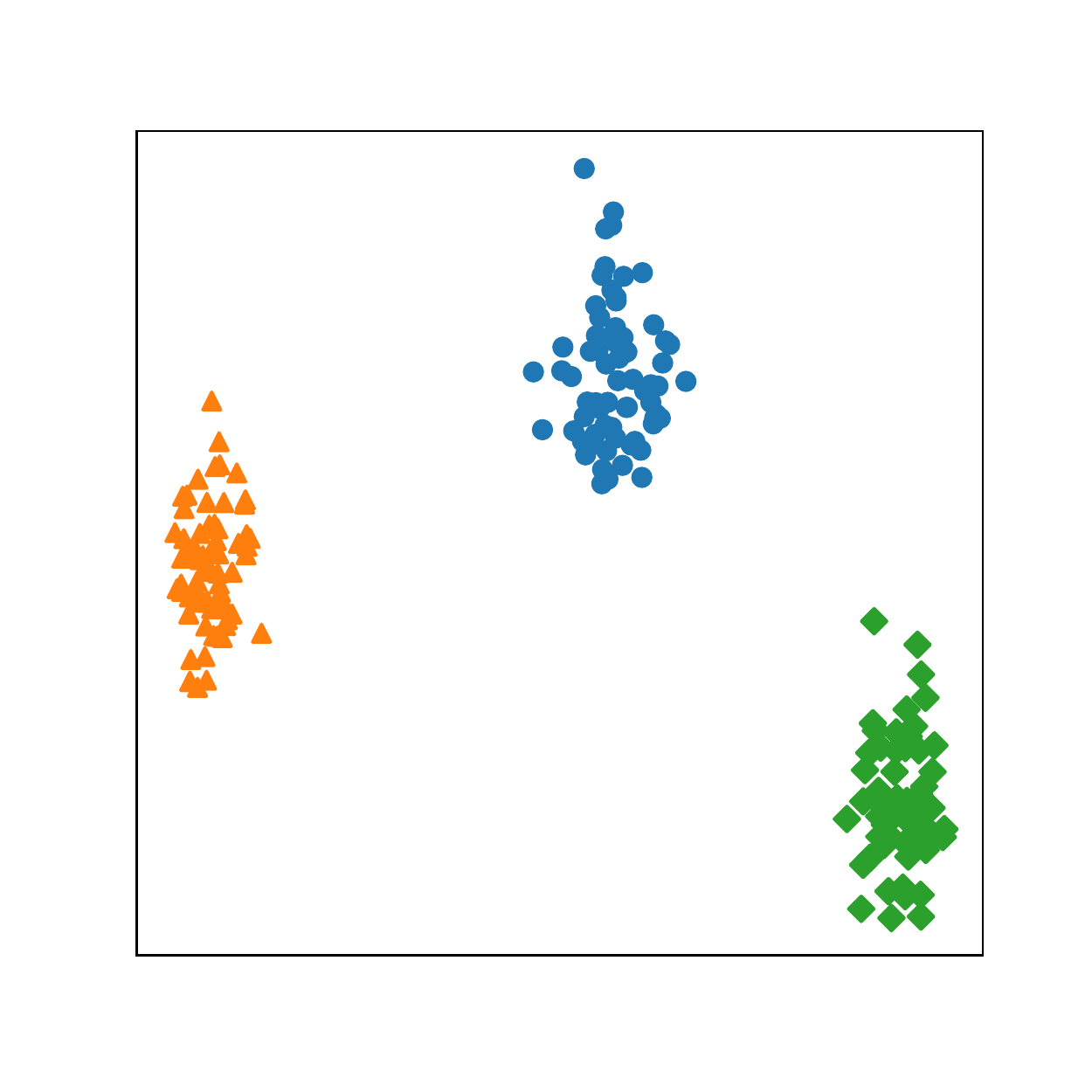}
         \caption{after drift ($M_d = 0.6, C_a = 3$)}
         \label{fig:aft_drft_3}
     \end{subfigure}
     \hfill
     \begin{subfigure}{0.24\textwidth}
         \centering
         \includegraphics[width=\textwidth]{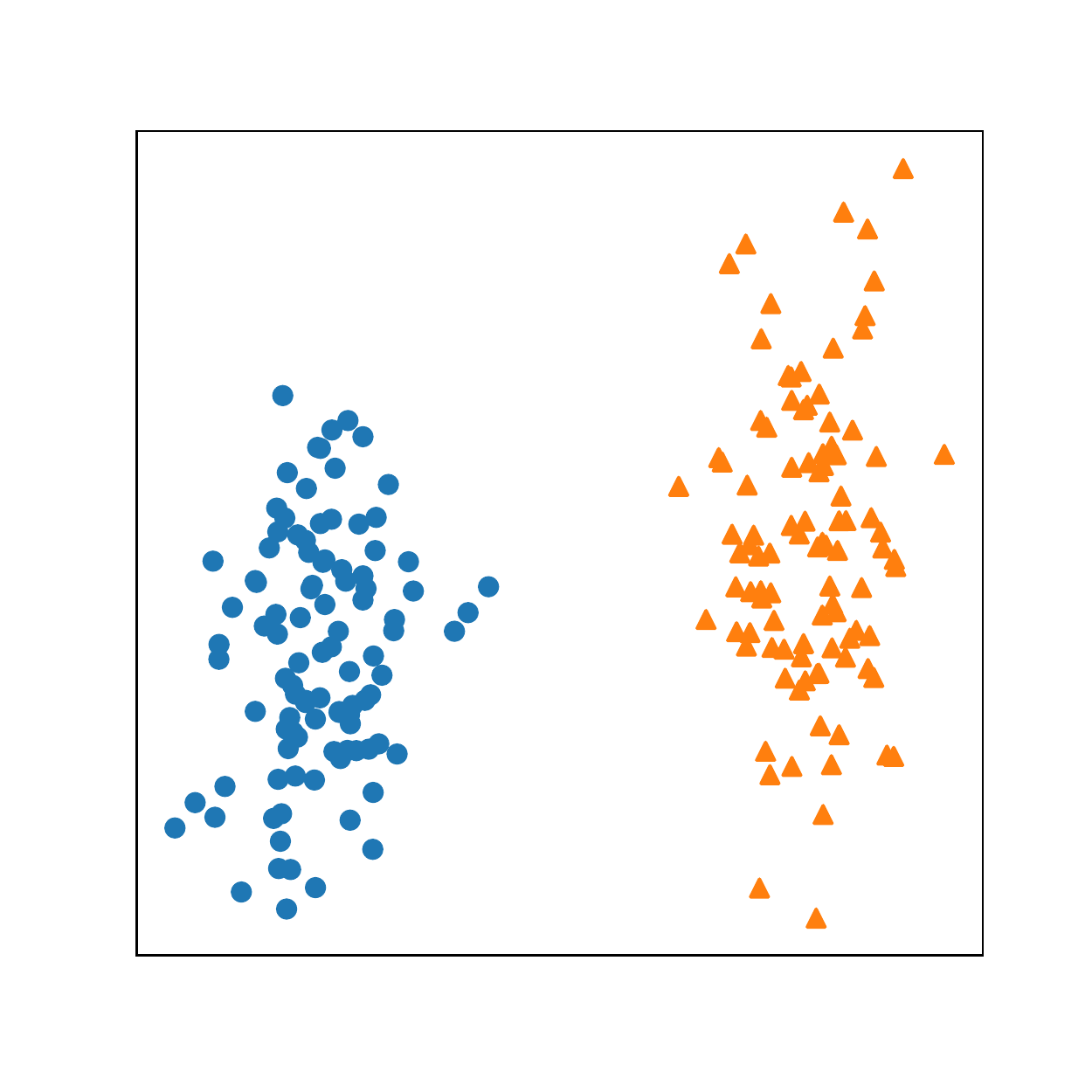}
         \caption{after drift ($M_d = 0.4, C_a = 2$)}
         \label{fig:aft_drft_4}
     \end{subfigure}
     \caption{Drifted Stream generation using \method{} ($M_d$: drift magnitude, $C_b/C_a$: \# clusters before/after drift) }
\end{figure*}

The mixture models generated by Algorithm \ref{alg:mdl_gen} can then used by the drifted-stream generator to synthesize a drifted data stream. The generator uses Algorithm \ref{alg:inst_gen} to create each instance of the stream. If the instance belongs to the pre-/post-drift concept, the generator samples from the pre-/post-drift mixture model respectively (lines 1-5). However, if the instance belongs to the time interval between the two concepts and the type of the drift is \textit{gradual}, the generator chooses one of the concepts randomly such that the probability of choosing the pre-/post-drift concept decreases/increases as we move towards the end of the interval (lines 6-12). In case of an \textit{incremental} drift, the generator assigns a weight to each mixture model based on the relative time gap between the instance and the pre- and post-drift concepts. Then, it uses that weight to determine the number of latent vectors to sample from each mixture model (lines 13-20). 

As an illustrative example, assume that we have 200 hosts and want to divide them into two clusters, Figure \ref{fig:bef_drft} shows the clusters generated by the drifted-stream generator before applying concept drift. The impact of an abrupt drift (i.e. $D_d = 0$) with a magnitude of $M_d = 0.6$ on these clusters is reflected in Figure \ref{fig:aft_drft}. Figure \ref{fig:aft_drft_3}  shows another case where the number of clusters changes after concept drift (i.e. we have concept evolution), and finally Figure \ref{fig:aft_drft_4} shows the case without any change in the number of clusters but with a lower drift magnitude.

\begin{algorithm}[tb]
\SetAlgoLined
\KwIn{ $Stream, Mix_{pre}, Mix_{pst}, T_d, D_d$}
$Size_s = |Stream|$\;
\uIf{$Size_s < N_b$}{
Inst = $Mix_{pre}$.sample($S_i$)\;
}
\uElseIf{$Size_s \geq N_b + D_d$}{
Inst = $Mix_{pst}$.sample($S_i$)\;
}
\uElseIf{$T_d$ = ''gradual''}{
Rnd = random(0,1)\;
\eIf{$Rnd < (Size_s - N_b)/D_d$}{
    Inst = $Mix_{pst}$.sample($S_i$)\;
}{
    Inst = $Mix_{pre}$.sample($S_i$)\;
}
}
\Else{
    $Weight = (Size_s - N_b)/D_d$\;
    $Size_{pre} = (1 - Weight) \times S_i$\;
    $Size_{pst} = Weight \times S_i$\;
    $Inst_{pre} = Mix_{pre}.sample(Size_{pre})$\;
    $Inst_{pst} = Mix_{pst}.sample(Size_{pst})$\;
    $Inst = Inst_{pre} \cup Inst_{pst}$\;
}
Stream.add(Inst)\;
 \caption{Instance Generation Algorithm}
 \label{alg:inst_gen}
\end{algorithm}

\begin{figure*}[t]
     \centering
     \captionsetup{justification=centering}
     \begin{subfigure}[b]{0.32\textwidth}
         \centering
         \includegraphics[width=\textwidth]{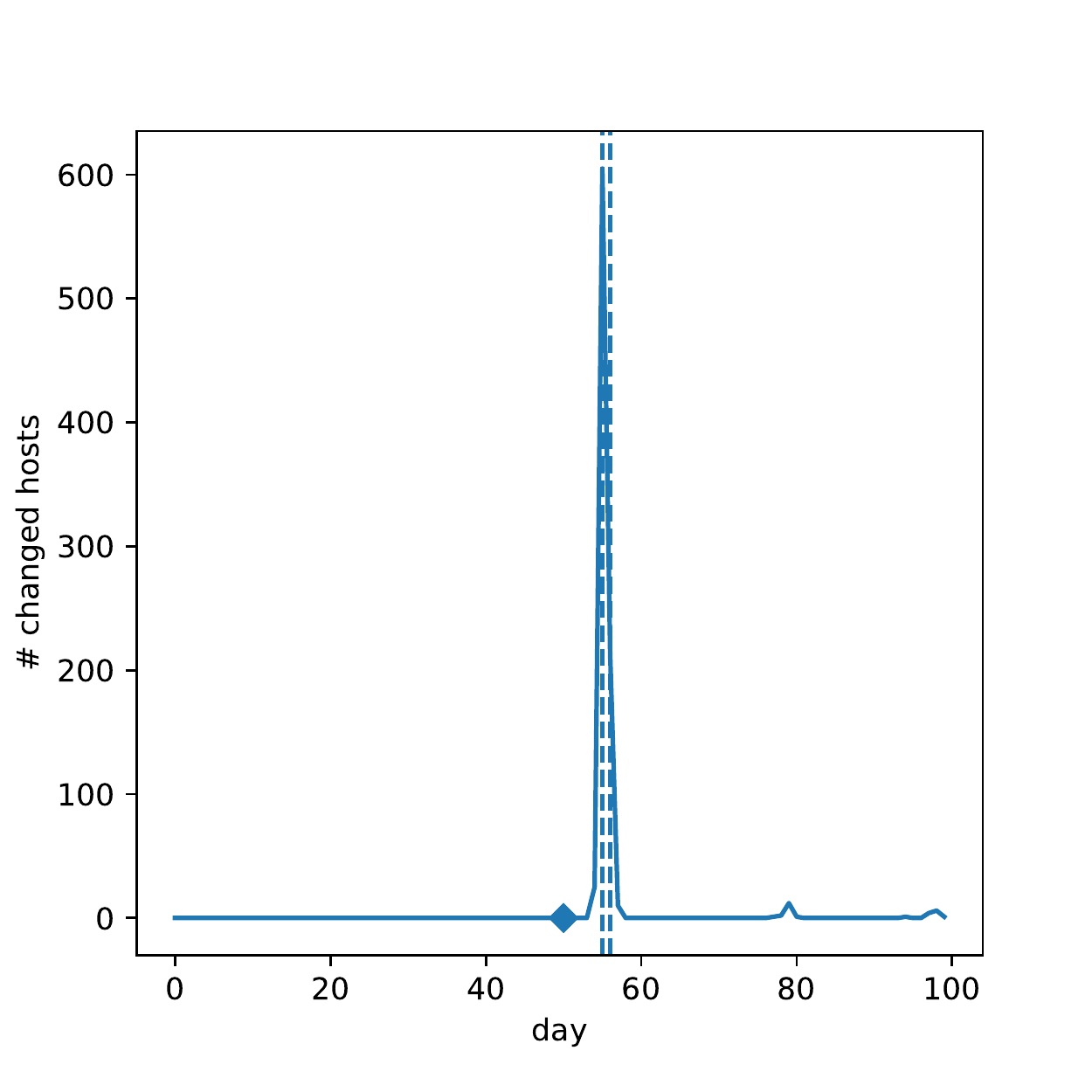}
         \caption{detection of abrupt drift  ($M_d = 0.6, C_b = C_a = 2$)}
         \label{fig:syn_a_det}
     \end{subfigure}
     \hfill
     \begin{subfigure}[b]{0.32\textwidth}
         \centering
         \includegraphics[width=\textwidth]{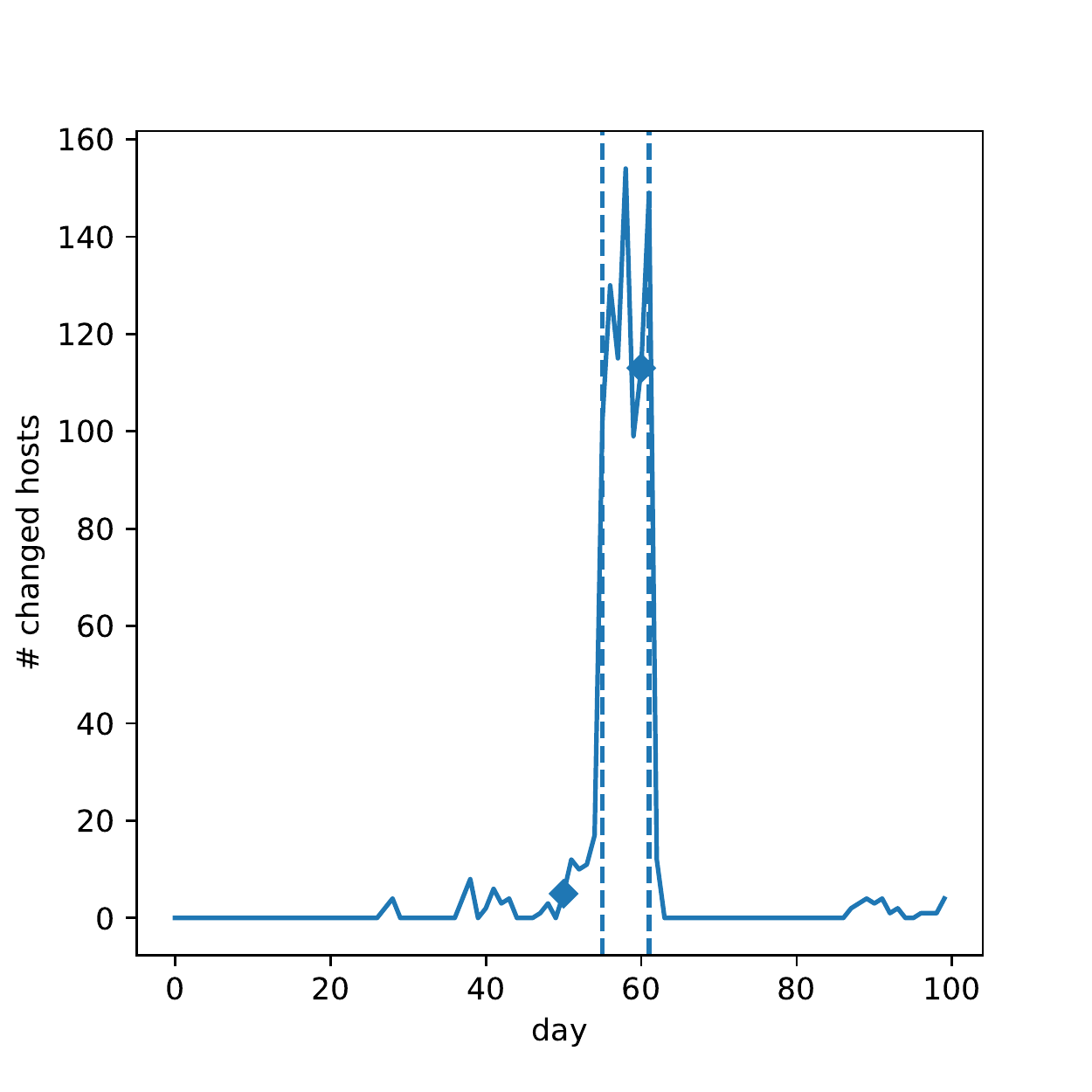}
         \caption{detection of gradual drift ($M_d = 0.6, C_b = C_a = 2, D_d = 10$)}
         \label{fig:syn_g_det}
     \end{subfigure}
     \hfill
     \begin{subfigure}[b]{0.32\textwidth}
         \centering
         \includegraphics[width=\textwidth]{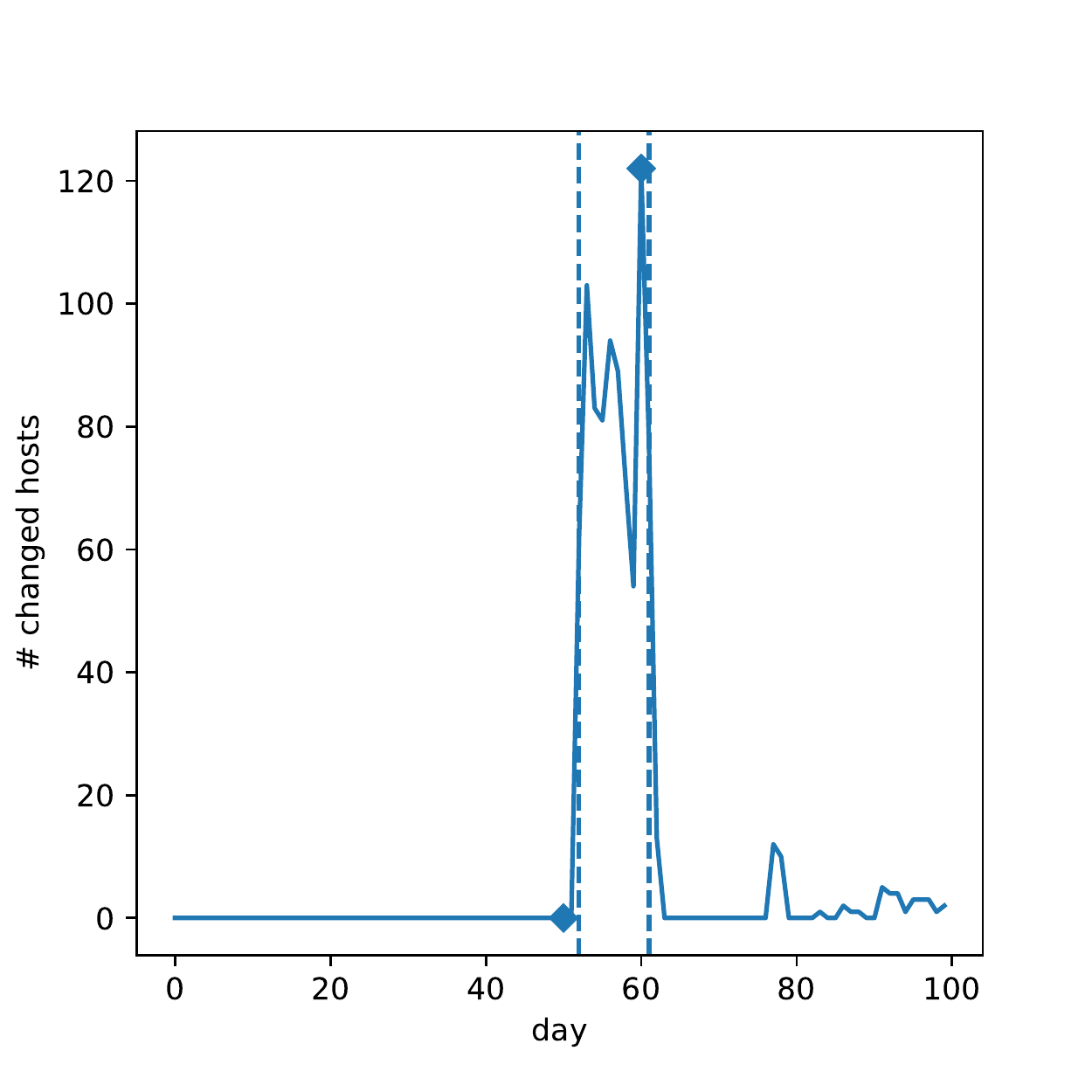}
         \caption{detection of incremental drift ($M_d = 0.6, C_b = C_a = 2, D_d = 10$)}
         \label{fig:syn_in_det}
     \end{subfigure}
     \caption{Detection of drift in the synthetic datasets ($M_d$: drift magnitude, $C_b/C_a$: \# clusters before/after drift, \\$D_d$: drift duration, diamond: start/end point of drift occurrence, dashed line: start/end point of drift detection)}
\end{figure*}

\begin{figure*}[b]
     \centering
     \captionsetup{justification=centering}
     \begin{subfigure}[b]{0.32\textwidth}
         \centering
         \includegraphics[width=\textwidth]{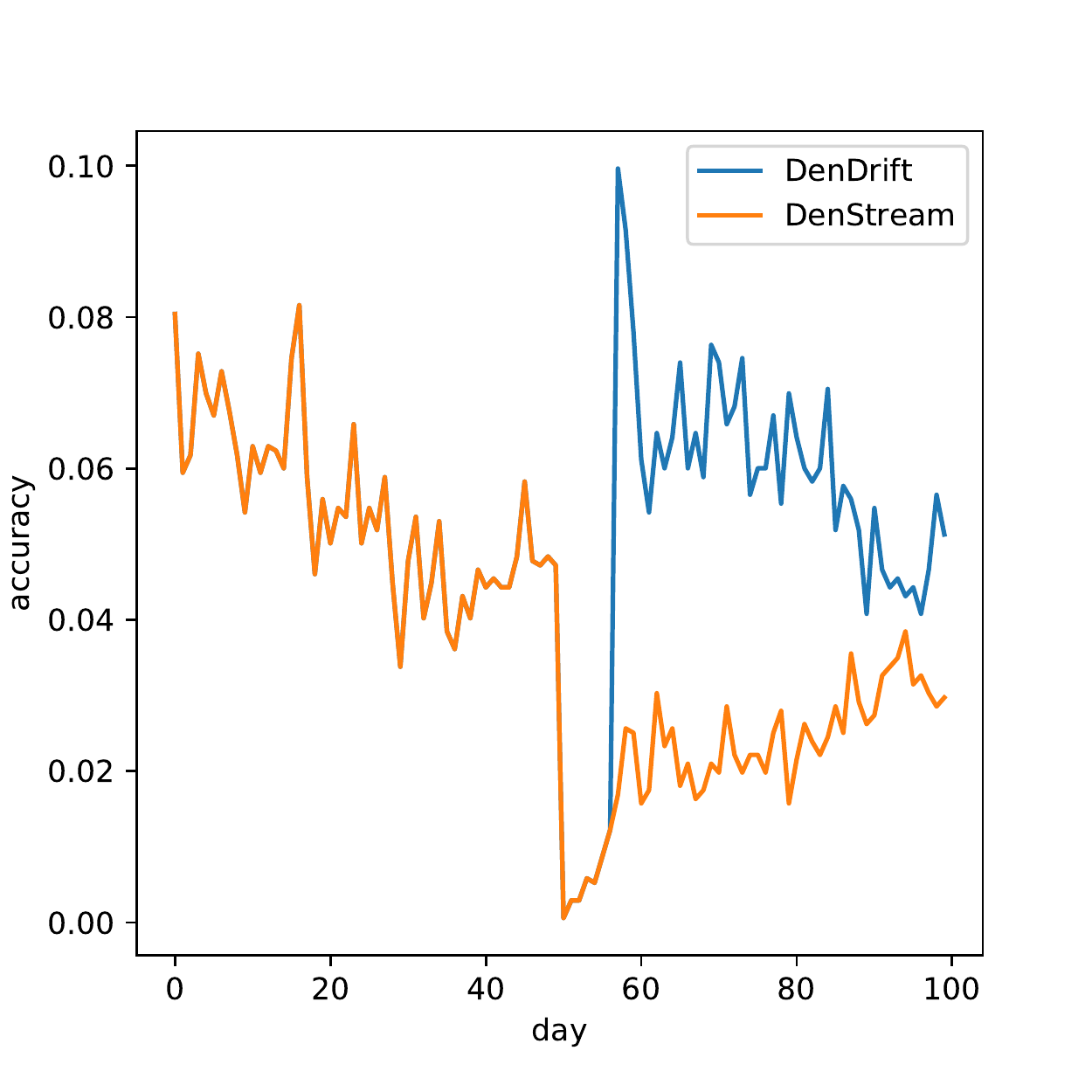}
         \caption{clustering accuracy (abrupt drift)}
         \label{fig:syn_a_v}
     \end{subfigure}
     \hfill
     \begin{subfigure}[b]{0.32\textwidth}
         \centering
         \includegraphics[width=\textwidth]{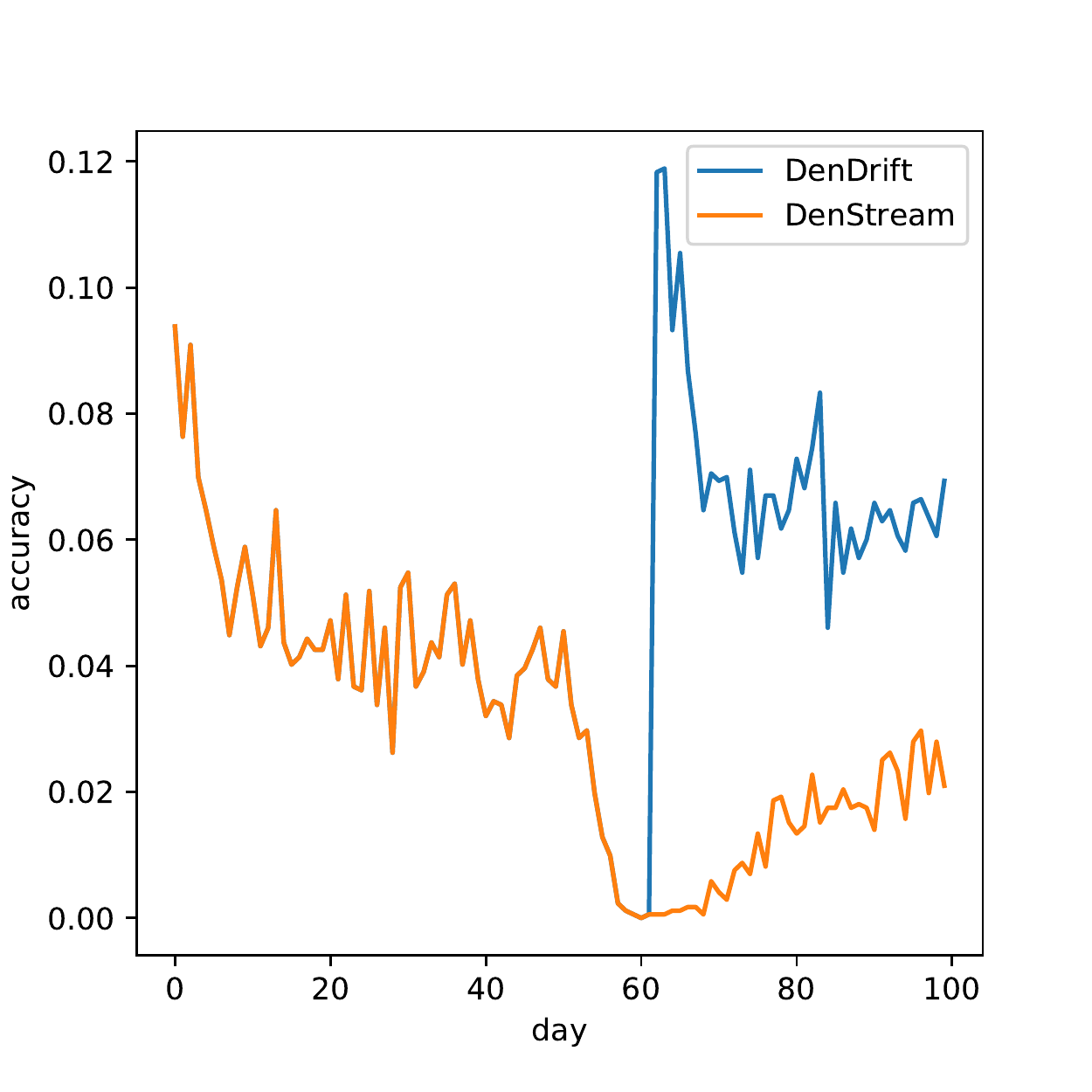}
         \caption{clustering accuracy (gradual drift)}
         \label{fig:syn_g_v}
     \end{subfigure}
     \hfill
     \begin{subfigure}[b]{0.32\textwidth}
         \centering
         \includegraphics[width=\textwidth]{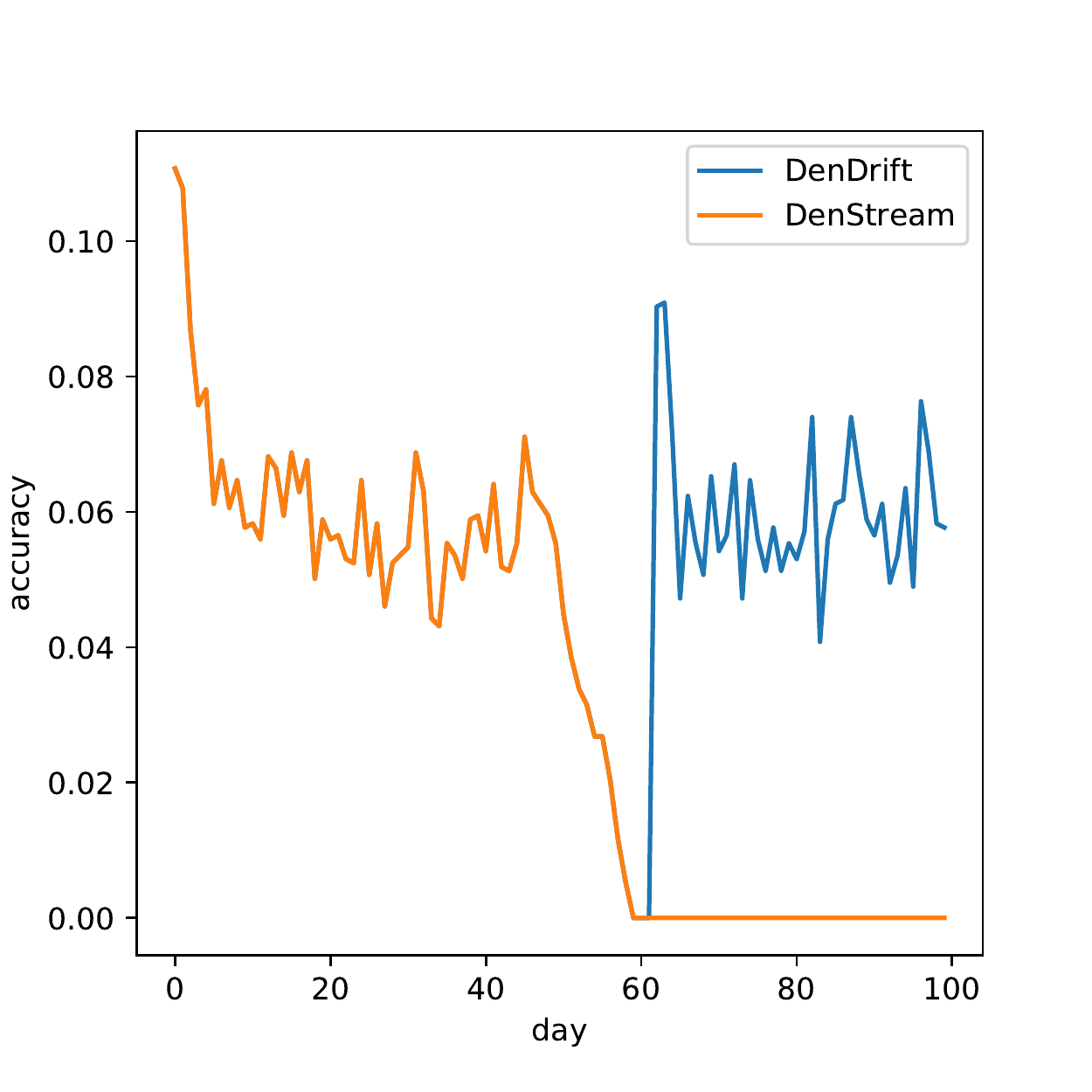}
         \caption{clustering accuracy (incremental drift)}
         \label{fig:syn_i_v}
     \end{subfigure}
     \caption{Quality evaluation of \method{} and DenStream on the synthetic dataset}
\end{figure*}

\subsection{Datasets}\label{sec:dataset}
We conducted experiments on both synthetic datasets generated using the drifted stream generator presented in Section \ref{sec:impl} and two datasets generated by applying NMF to the event log provided by F-Secure.

\subsubsection{F-Secure Datasets} \hfill \break
The F-Secure event log includes events recorded over a 3-month period (i.e. 92 days). Each record is a tuple of the form  (\textit{host\_id}, \textit{process\_name}, \textit{date}, \textit{count}), where \textit{count} indicates the number of times the process with identifier \textit{process\_name} (hashed for rare processes) is executed by the host with hashed identifier \textit{host\_id} on the date specified by \textit{date}. This log file includes data recorded for approximately 2000 hosts and 1,000,000 processes.
We partitioned this log file by date, such that each partition included data recorded for a single day. Then we applied NMF to each partition to obtain the host latent vectors for the corresponding day. To be able to analyze the potential influence of the number of latent features on the performance of \method{}, We considered two cases regarding the number of latent features for each host. One case with two and another with 10 latent features. At the end, we had two datasets consisting of 92 data instances. Each instance was a matrix with $N_h$ rows and two or 10 columns such that $N_h \approx 2000$ denotes the number of hosts. These datasets were label-less. To facilitate  comparing the quality of clusters obtained by \method{}  with those found by DenStream, we utilized the inherent clustering property of NMF \cite{ding2005equivalence} and assigned the index of the feature with the maximum value as a class label to each row of these datasets. 

\subsubsection{Synthetic Datasets}\hfill \break
We generated three synthetic datasets corresponding to the three forms of concept drift investigated in this paper (i.e. abrupt, gradual and incremental). Each of the synthetic datasets includes 100 instances with the same number of rows as the F-Secure datasets, and with two columns. In contrast to the F-Secure datasets, we know exactly the location, type and duration of concept drifts in the synthetic datasets. Furthermore, we have access to the ground truth (i.e. the correct label of each row), since we know which multi-variate normal distribution is used to generate each row.
For all synthetic datasets, we considered two as the number of clusters before concept drift ($C_b = 2$), and we inserted a concept drift with magnitude $M_d = 0.6$ in the 50th instance, such that the duration of gradual and incremental drifts was set to 10 ($D_d = 10$), and the minimum number of changed hosts to consider as a drift was 50 meaning that $Th_d \approx 0.03$.

\begin{figure*}[t]
     \centering
     \captionsetup{justification=centering}
     \begin{subfigure}[b]{0.32\textwidth}
         \centering
         \includegraphics[width=\textwidth]{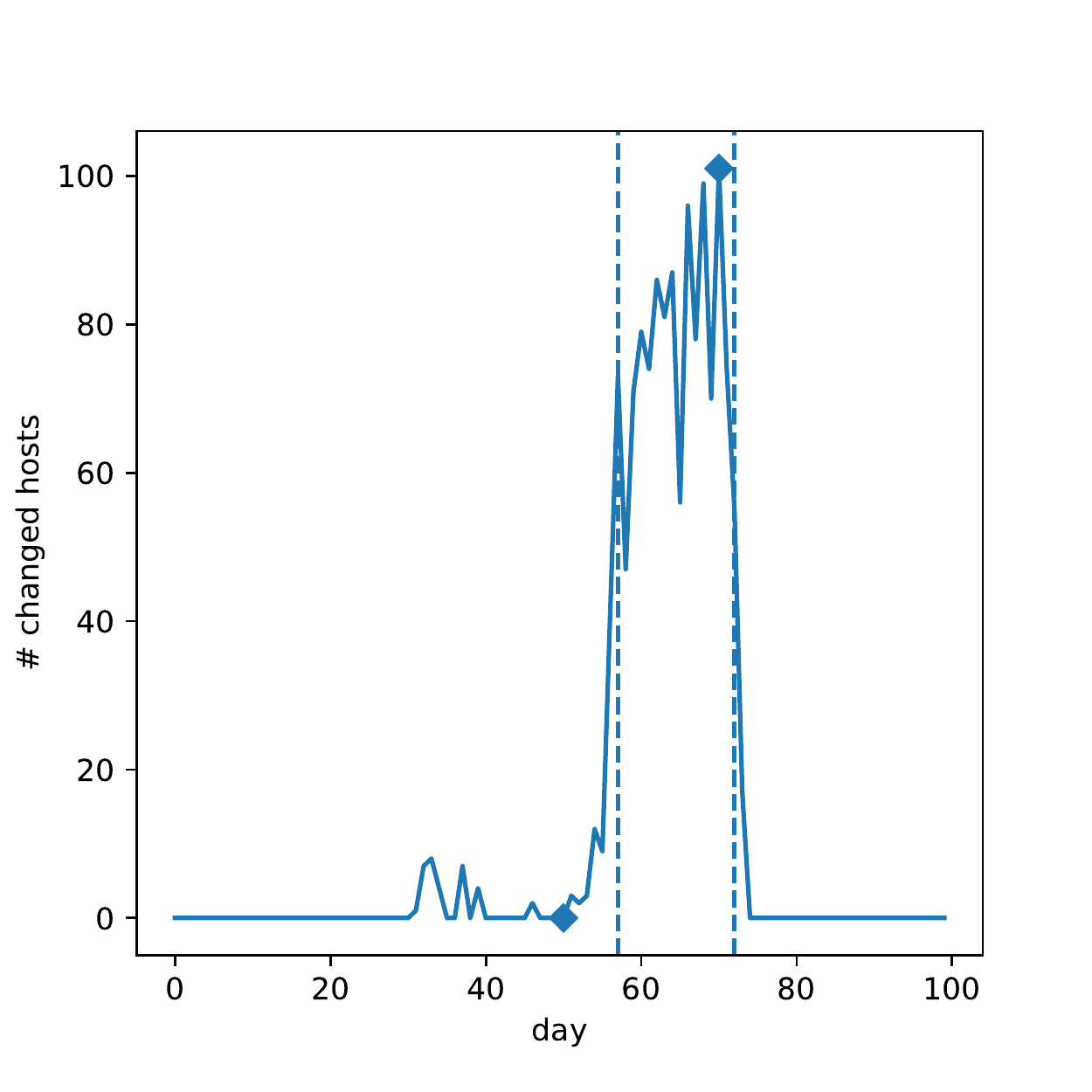}
         \caption{detection of drifts with concept evolution (($M_d = 0.6, C_b = 2, C_a = 4, D_d = 20$))}
         \label{fig:evolv_det}
     \end{subfigure}
     \hfill
     \begin{subfigure}[b]{0.32\textwidth}
         \centering
         \includegraphics[width=\textwidth]{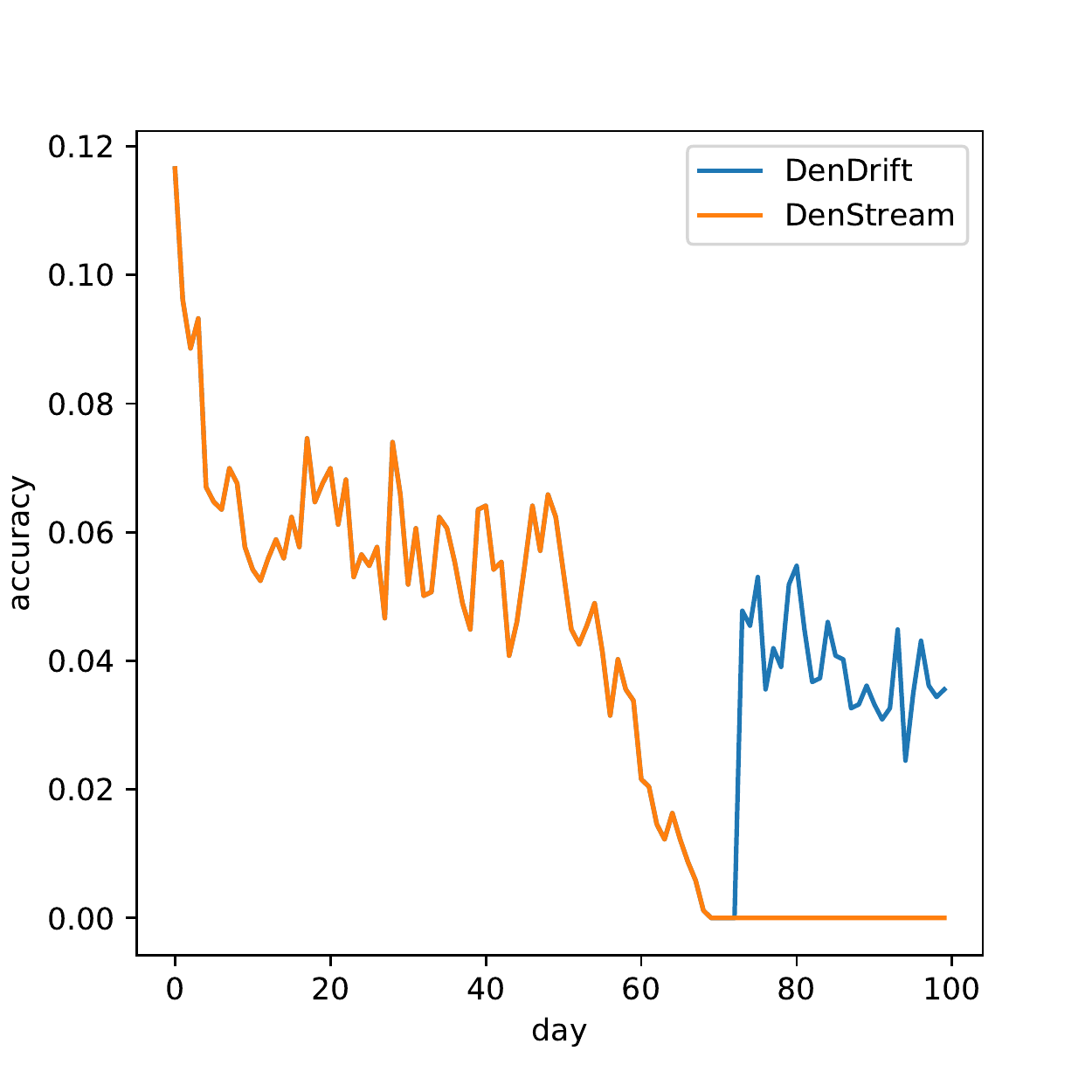}
         \caption{clustering accuracy ($M_d = 0.6, C_b = 2, C_a = 4, D_d = 20$) }
         \label{fig:grow_acc}
     \end{subfigure}
     \hfill
     \begin{subfigure}[b]{0.32\textwidth}
         \centering
         \includegraphics[width=\textwidth]{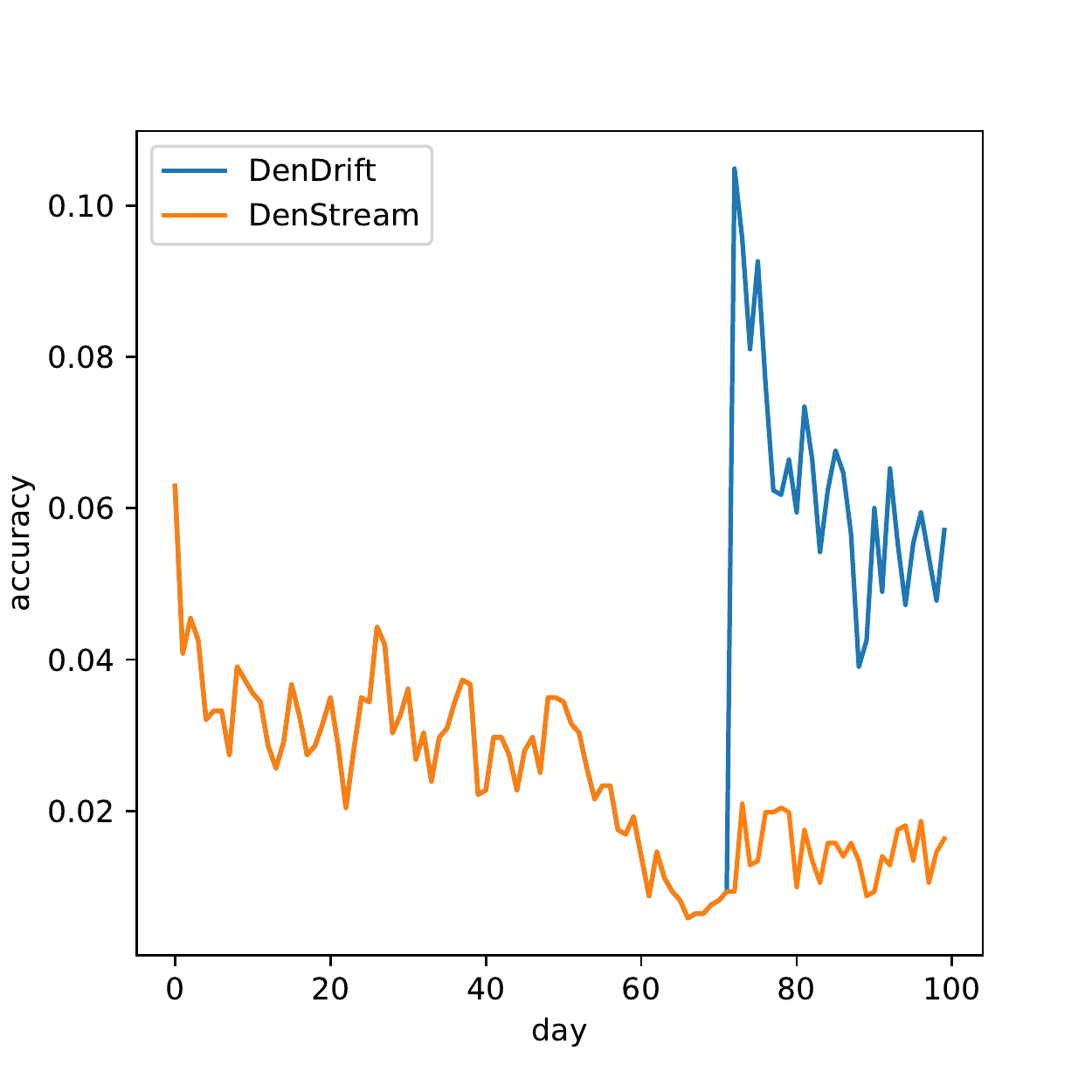}
         \caption{clustering accuracy ($M_d = 0.6, C_b = 4, C_a = 2, D_d = 20$) }
         \label{fig:fall_acc}
     \end{subfigure}
     \caption{\method{} and concept evolution ($M_d$: drift magnitude, $C_b/C_a$: \# clusters before/after drift, $D_d$: drift duration)}
     \label{fig:evolv}
\end{figure*}

\begin{figure*}[b]
     \centering
     \begin{subfigure}[b]{0.24\textwidth}
         \centering
         \includegraphics[width=\textwidth]{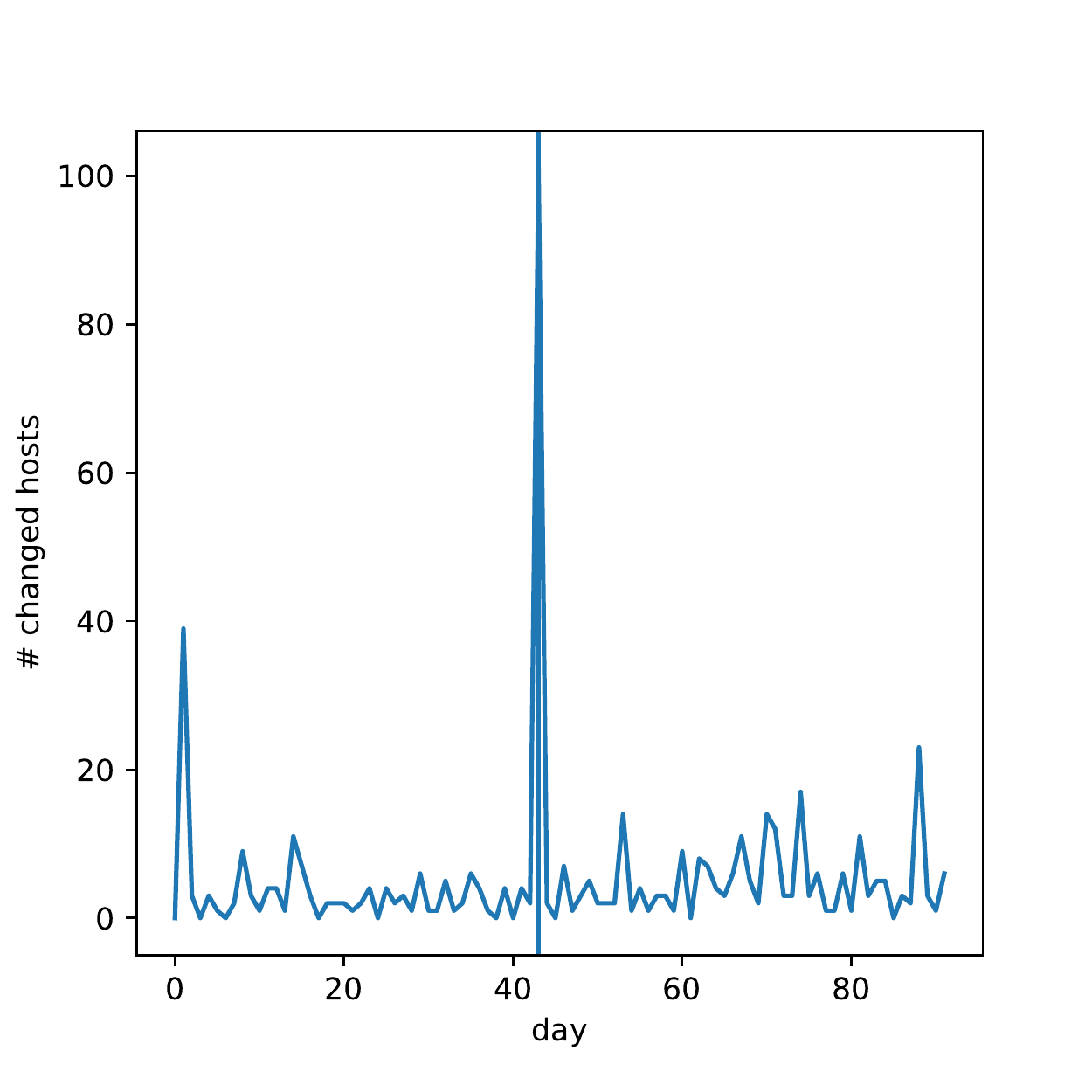}
         \caption{drift detection ($N_f = 2$)}
         \label{fig:fsec_det_2}
     \end{subfigure}
     \hfill
     \begin{subfigure}[b]{0.24\textwidth}
         \centering
         \includegraphics[width=\textwidth]{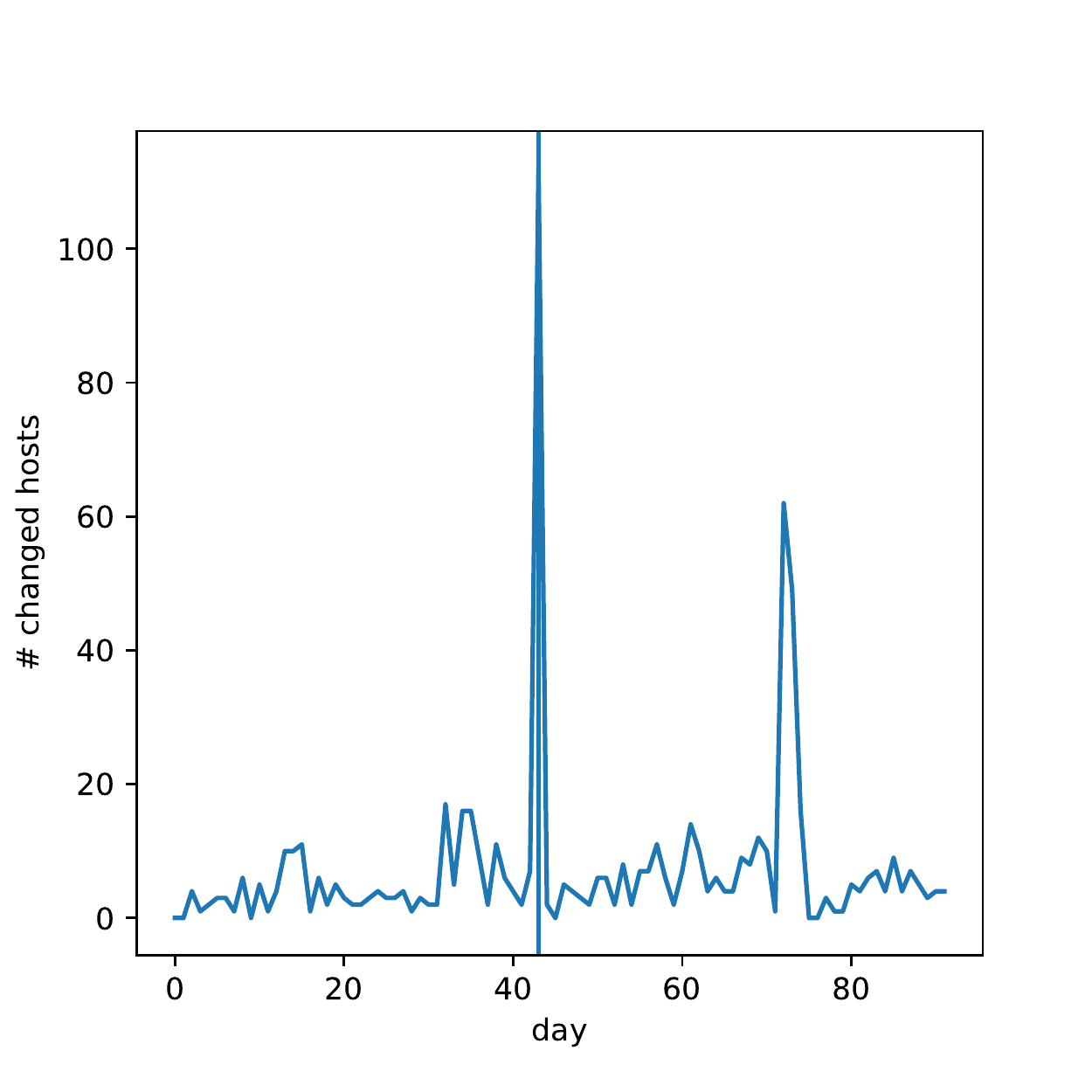}
         \caption{drift detection ($N_f = 10$)}
         \label{fig:fsec_det_10}
     \end{subfigure}
     \hfill
     \begin{subfigure}[b]{0.24\textwidth}
         \centering
         \includegraphics[width=\textwidth]{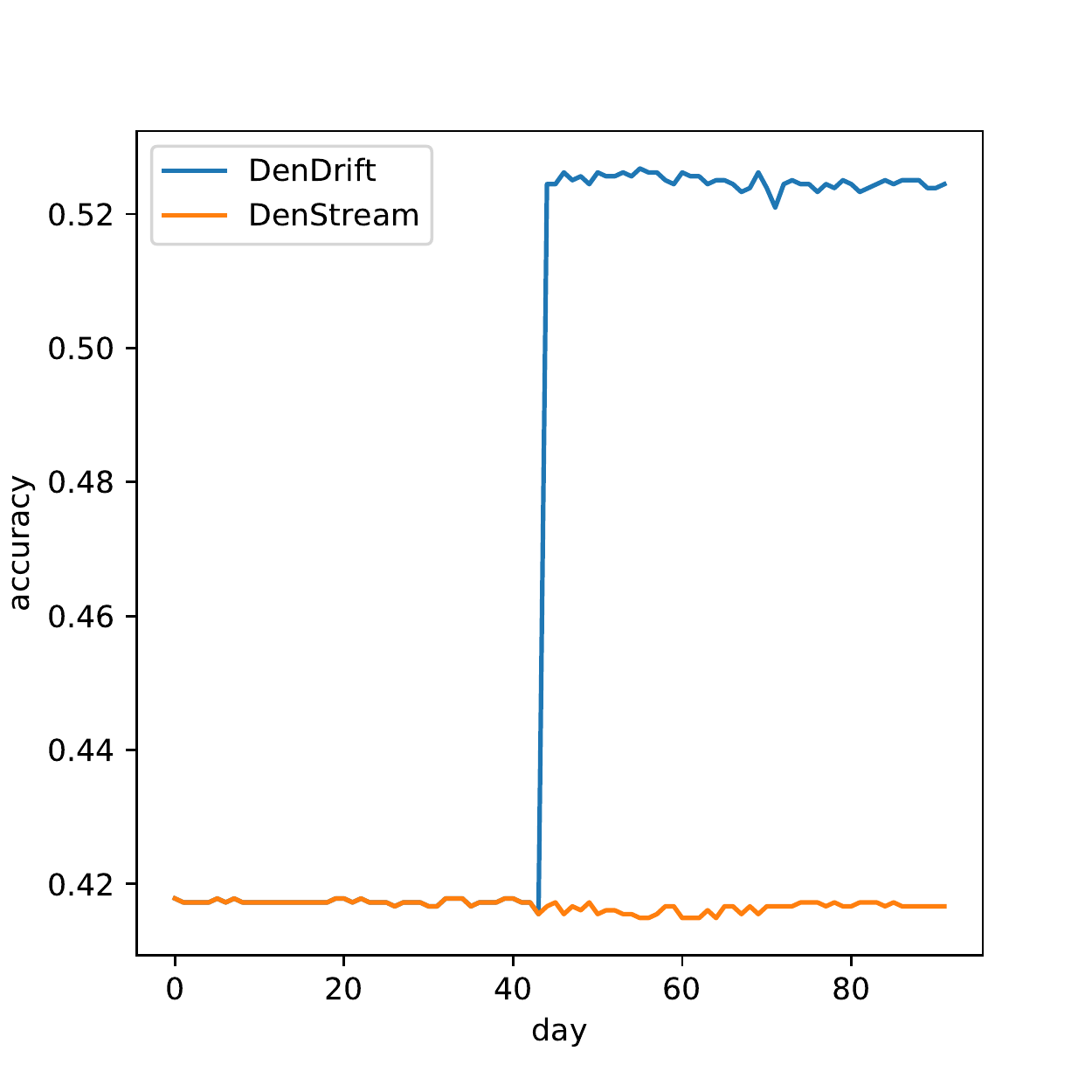}
         \caption{quality comparison ($N_f = 2$)}
         \label{fig:fsec_acc_2}
     \end{subfigure}
     \hfill
     \begin{subfigure}[b]{0.24\textwidth}
         \centering
         \includegraphics[width=\textwidth]{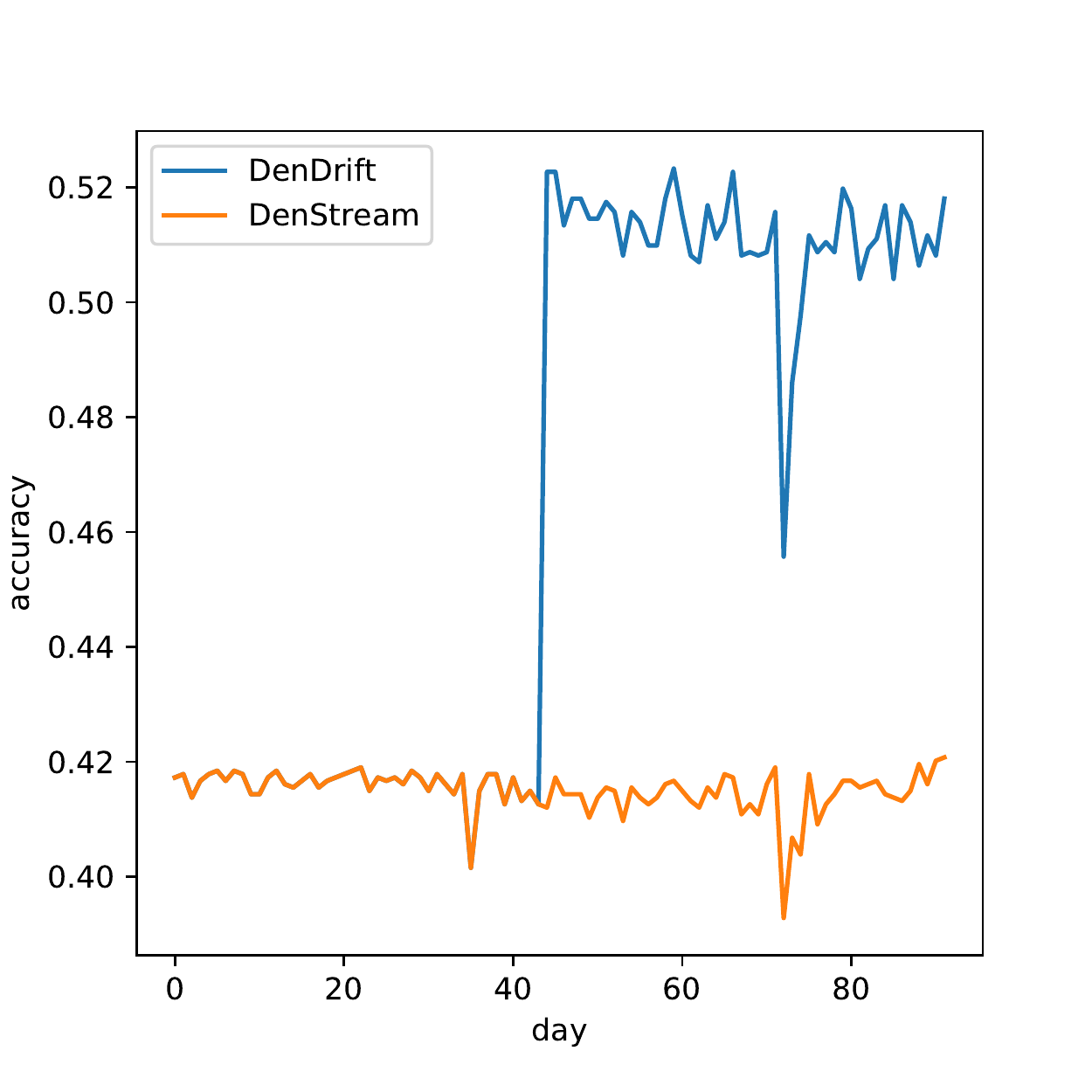}
         \caption{quality comparison ($N_f = 10$)}
         \label{fig:fsec_acc_10}
     \end{subfigure}
     \caption{Performance of \method{}  on the industrial dataset}\label{fig:fsec}
\end{figure*}

\subsection{Observations}\label{sec:obs}
\subsubsection{Synthetic Datasets}\hfill \break
Figures \ref{fig:syn_a_det}-\ref{fig:syn_in_det} show the number of changes detected by \method{} in each synthetic dataset. 
The start and end points of concept drift are highlighted by a diamond, and the start/end points detected by \method{} are indicated be a dashed line. According to these results, \method{} can detect the three forms of concept drift with a short delay. In these experiments, we configured DenStream with $\lambda = 0.1$, $\epsilon = 0.5$, $\beta = 0.8$ and $\mu = 3$. The impact of retraining (after drift detection) on the accuracy of clustering is demonstrated in Figures \ref{fig:syn_a_v}-\ref{fig:syn_i_v}. Accordingly, compared to DenStream, \method{}  is more robust against concept drift. In particular, we see a considerable degradation in the accuracy of DenStream due to concept drift. Even in some cases, (e.g. Figure \ref{fig:syn_i_v}), the accuracy of DenStream stays very low after concept drift. On the contrary, \method{}  can recover quickly.

In the above experiments, we assumed that the number of clusters would not change due to concept drift. However, we also did experiments on cases with evolving concepts, where we had a different number of clusters before and after drift. Since the drift detection mechanism in \method{}  is independent of the number of clusters and clustering mechanism, any change in the number of clusters would not have a negative impact on the detection performance of \method{}. As shown in Figure \ref{fig:evolv_det}, \method{}  can effectively detect concept drift in spite of concept evolution. In this example, we have a gradual drift with a duration of 20 ($D_d = 20$) such that the number of cluster increases from two to four after the concept drift. Furthermore, Figures \ref{fig:grow_acc} and \ref{fig:fall_acc} compare the accuracy of \method{}  with DenStream in cases that we have an increase or decrease in the number of clusters respectively. In both cases, \method{} has a better performance.

\subsubsection{F-Secure Datasets}\hfill \break
Figure \ref{fig:fsec} reports the results of the experiments performed on the F-Secure datasets. Accordingly, \method{}  has detected an abrupt drift on day 43 in both cases. In particular, more than 100 hosts have found to be changed on this day irrespective of the number of latent features. We found that this change was related to a pilot testing which was started a few days before and led to a change in the behavior of the employed hosts. As Figures \ref{fig:fsec_acc_2}-\ref{fig:fsec_acc_10} demonstrate, this change did not have a significant negative impact on the quality of clusters. However, retraining DenStream for the new concept, improved its accuracy by approximately 10\%. Another fact worth mentioning is the outlier detected on day 72 for the dataset with 10 latent features (Figure \ref{fig:fsec_det_10}). In particular, \method{}  detected a sudden change in more than 150 hosts. However, this change was not a true concept drift rather a temporary change (or an outlier). Since a temporary change does not change the concept (statistical properties), it is not necessary to retrain DenStream. In these experiments, we configured DenStream with $\lambda = \epsilon = 0.1$, $\beta = 1$ and $\mu = 9$.

\begin{figure}[t]
    \centering
    \includegraphics[width=0.35\textwidth]{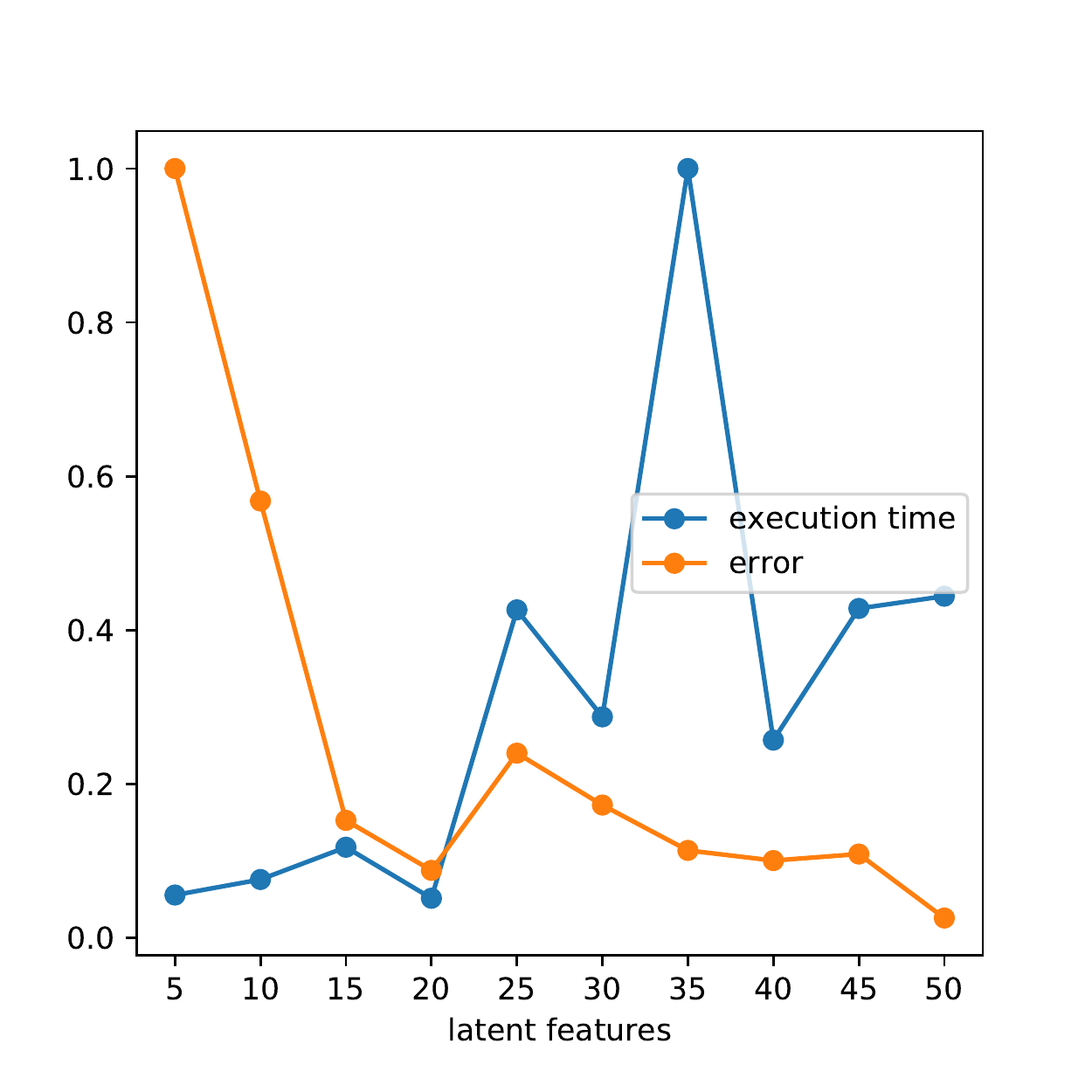}
    \caption{Reconstruction error vs. execution time for NMF}
    \label{fig:time_err}
\end{figure}

\begin{figure}[tb]
    \centering
    \includegraphics[width=0.35\textwidth]{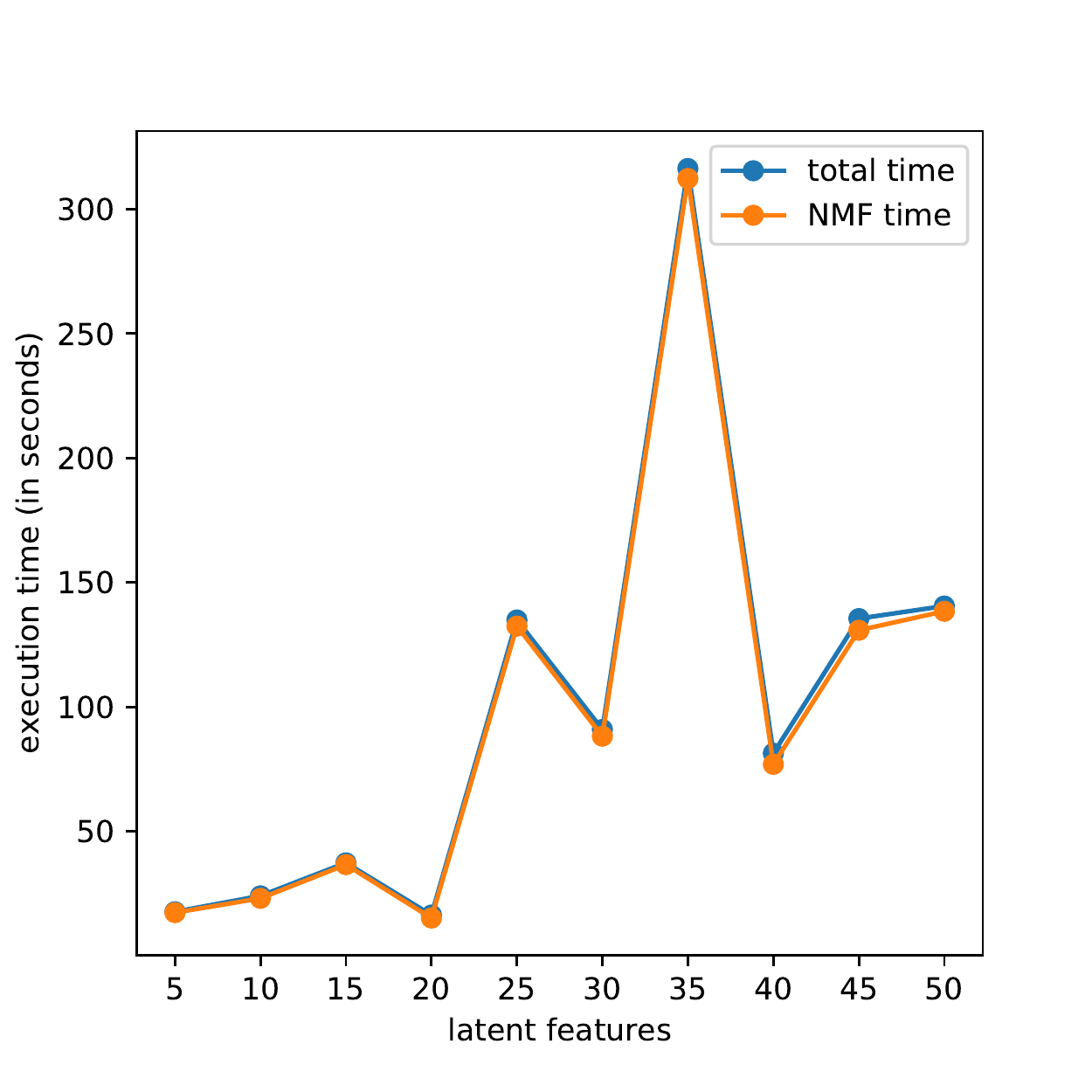}
    \caption{Execution time vs dimensionality}
    \label{fig:effort}
\end{figure}

The results presented in Figure \ref{fig:fsec} highlight that the number of changed hosts may potentially differ depending on the number of latent features (i.e. $N_f$), and the thresholds for change and drift (i.e. $Th_c$ and $Th_d$ in Algorithm \ref{alg:\method{}}. Apparently, if we have very low thresholds, \method{} may enter the \textit{change} mode more frequently. However, the capability of \method{} to distinguish drifts from outliers, would avoid false positives and unnecessary retrainings in most cases. Another fact worth discussion is the impact of $N_f$ on the efficiency and accuracy of NMF and hence \method{}. If we pick a too small/large value, we would end up with underfitting/overfitting in the NMF model.  We did an experiment on NMF regarding the relation between $N_f$ and both the efficiency and accuracy of factorization. In this experiment, we considered 200 as the maximum number of iterations until convergence. We observed that increasing the number of latent features would decrease reconstruction error (or increase factorization accuracy) and may also increase the execution overhead of NMF (see Figure \ref{fig:time_err}). Reconstruction error measures the difference between the input host-process matrix $M$ and $W \times H$, where $W$ and $H$ are the matrices generated by NMF in \method{}  (line 3 in Algorithm \ref{alg:\method{}}). Figure \ref{fig:time_err} indicates that 20 is an optimal choice for the F-Secure data if we take into account only the efficiency and accuracy of NMF. 
We performed another experiment to assess the impact of $N_f$ on the overall execution overhead of \method{}. Figure \ref{fig:effort} shows the relation between $N_f$ and both the overhead of NMF and the total execution time of \method{} for processing one host-process matrix. Accordingly, in all cases the time spent for drift detection and clustering is negligible in comparison to the delay imposed by NMF, and the NMF computations may take more than five minutes due to the delay in convergence.   

\section{Concluding Remarks}\label{sec:conclusion}
In this paper, we presented \method{}  as a drift-aware algorithm for host profiling. \method{}  is an extension of an incremental clustering algorithm called DenStream and relies on Page-Hinckley test for concept drift detection. Furthermore, we have developed a concept drift simulator which can generate drifted streams with a user-defined dimensionality, duration and drift magnitude. Three prevalent forms of concept drift are supported by this simulator (i.e. abrupt, gradual and incremental), which can be injected at a user-defined point in the stream. Our experimental results indicate that \method{}  can detect and adapt to abrupt, gradual and incremental drifts in synthetic streams generated by our simulator. Furthermore, our experiments on real-world data with an abrupt drift also confirm the effectiveness of \method{}  for drift detection and adaptation. Furthermore, our experiments on high-dimensional streams indicate that \method{} can effectively distinguish between drifts and outliers (or temporary changes), and avoid unnecessary retraining. 
\method{} is principally an extension of DenStream, but the drift detection mechanism is independent of the clustering mechanism. Therefore, it is easy to use other algorithms (e.g. CluStream and StreamKM++) as the underlying clustering algorithm in \method{}. Doing further experiments on those algorithms is considered as future work.

\begin{acks}
This work was carried out during the tenure of an ERCIM ‘Alain Bensoussan’ Fellowship Program and has been partially supported and funded by the ITEA3 European \href{https://itea3.org/project/ivves.html}{IVVES} project. 
\end{acks}

\bibliographystyle{ACM-Reference-Format}
\bibliography{refs}


\end{document}